\documentclass[10pt,journal,compsoc,twocolumn ]{IEEEtran}

\usepackage{times}
\usepackage{epsfig}
\usepackage{graphicx}
\usepackage{amsmath}
\usepackage{amssymb}
\usepackage{multirow}
\usepackage{bbm}
\usepackage{rotating}
\usepackage{array}
\usepackage{threeparttable}

\usepackage{epstopdf}
\usepackage{subfigure}
\usepackage{amsmath}
\usepackage{bm}
\usepackage{url}
\usepackage{algorithm}
\usepackage{algorithmic}
\usepackage[none]{hyphenat}

\begin{document}

\title{Adaptive Face Recognition Using Adversarial Information Network}

\author{Mei Wang, Weihong Deng
\thanks{The authors are with the Pattern Recognition and Intelligent System Laboratory, School of Artificial Intelligence, Beijing University of Posts and Telecommunications, Beijing, 100876, China. E-mail: \{wangmei1,whdeng\}@bupt.edu.cn. (Corresponding author: Weihong Deng)}}

\maketitle

\begin{abstract}

In many real-world applications, face recognition models often degenerate when training data (referred to as source domain) are different from testing data (referred to as target domain). To alleviate this mismatch caused by some factors like pose and skin tone, the utilization of pseudo-labels generated by clustering algorithms is an effective way in unsupervised domain adaptation. However, they always miss some hard positive samples. Supervision on pseudo-labeled samples attracts them towards their prototypes and would cause an \emph{intra-domain gap} between pseudo-labeled samples and the remaining unlabeled samples within target domain, which results in the lack of discrimination in face recognition. In this paper, considering the particularity of face recognition, we propose a novel adversarial information network (AIN) to address it. First, a novel adversarial mutual information (MI) loss is proposed to alternately minimize MI with respect to the target classifier and maximize MI with respect to the feature extractor. By this min-max manner, the positions of target prototypes are adaptively modified which makes unlabeled images clustered more easily such that intra-domain gap can be mitigated. Second, to assist adversarial MI loss, we utilize a graph convolution network to predict linkage likelihoods between target data and generate pseudo-labels. It leverages valuable information in the context of nodes and can achieve more reliable results. The proposed method is evaluated under two scenarios, i.e., domain adaptation across poses and image conditions, and domain adaptation across faces with different skin tones. Extensive experiments show that AIN successfully improves cross-domain generalization and offers a new state-of-the-art on RFW dataset.

\end{abstract}


\section{Introduction}

Recently, great progress has been achieved in face recognition (FR) \cite{wang2018deep} with the help of large-scale datasets and deep convolutional neural networks (CNN) \cite{krizhevsky2012imagenet,simonyan2014very,he2016deep}. 
However, FR models trained with the assumption that training data are drawn from similar distributions with testing data often degenerate seriously due to the mismatch of distribution caused by poses and skin tones of faces. For instance, FR systems, trained on existing databases where frontal faces appear more frequently, demonstrate low recognition accuracy when applied to the images with large poses. Re-collecting and re-annotating a large-scale training dataset for each specific testing scenario is effective but extremely time-consuming. The concern about privacy also makes it difficult to get access to labeled images for training. Therefore, how to guarantee the generalization ability for automatic systems and prevent the performance drop in real-world applications is realistic and meaningful, but few works have focused on it.

\begin{figure}
\centering
\includegraphics[width=9cm]{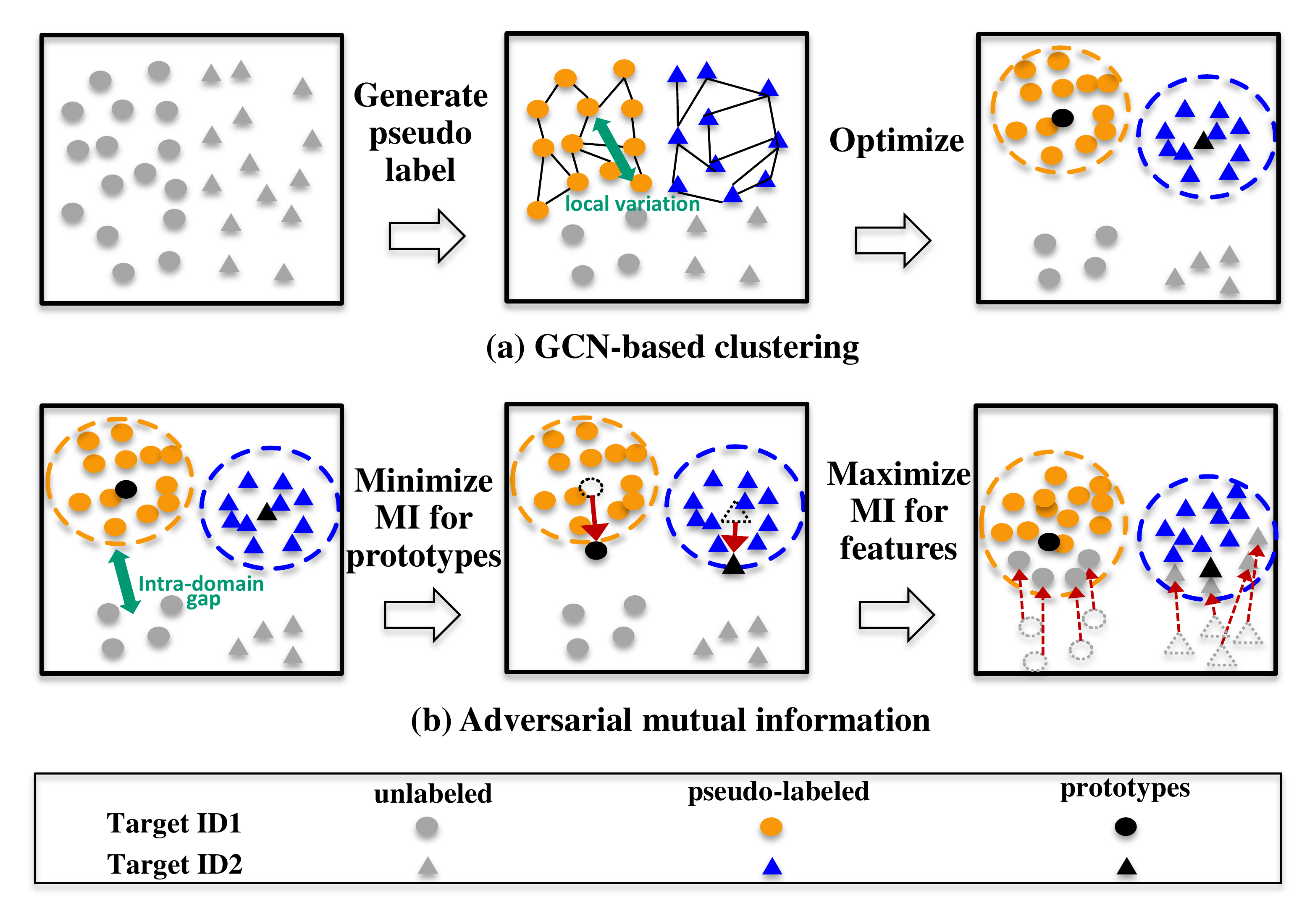}
\caption{Our AIN addresses intra-domain gap within target domain leading to discriminative feature space. (a) GCN can leverage the relationships between node neighbors to generate pseudo-labels by which the source model is adapted and variations can be reduced.  (b) Supervision on pseudo-labeled samples attracts the corresponding features towards their prototypes, which results in intra-domain gap between pseudo-labeled and unlabeled samples. Our adversarial MI learning can mitigate this gap in a min-max manner.}
\label{general} 
\end{figure}

In object classification, unsupervised domain adaptation (UDA) \cite{wang2018deepdomain} is proposed to overcome domain mismatch. UDA can learn transferable features using a labeled source domain and an unlabeled target domain, such that models trained on source domain will perform well on target domain. However, UDA in face recognition is more realistic but challenging since source and target domains have completely disjoint classes, which is even stricter than the assumption of open-set domain adaptation \cite{panareda2017open} in object classification. Due to the particularity of FR, traditional UDA methods are usually unsuitable.

Popular UDA methods by domain alignment \cite{Long2015Learning,Tzeng2017Adversarial} only consider the inter-domain shift, but largely overlook the variations within target domain which are the keys for learning discriminative features in FR model. Adapting networks with pseudo-labels is an effective way to reduce these variations, but clustering-based pseudo-labels \cite{Kang_2019_CVPR,wang2019racial} are usually unreliable in FR. Thus, some recent methods \cite{zhang2018collaborative,chen2018progressive} proposed to only assign pseudo-labels for the target samples with high confidence, but they discarded useful information encoded in the remaining unlabeled samples. 
Moreover, these methods tend to separate the target distribution into two subdistributions, thus causing the intra-domain gap as shown in Fig. \ref{general}-(b). The presence of pseudo-labels pulls these pseudo-labeled samples towards their feature clusters. Besides, the remaining unlabeled samples which less correlate with the given pseudo-labeled samples are located far from these clusters and thus are more difficult to be attracted to them even if unlabeled data are optimized by unsupervised losses \cite{wang2019racial,mei2020instance}. 
Therefore, it is essential for domain adaption to enhance the clustering quality and avoid the separation of intra-class samples in the target domain.

In this paper, we propose an adversarial information network (AIN), which explicitly considers intra-domain gap of target domain and learns discriminative target distribution under an unsupervised setting to improve cross-domain generalization in FR. First, graph convolutional network (GCN) \cite{bruna2013spectral,kipf2016semi} is utilized to predict positive neighbors in target domain and generate pseudo-labels which can help to reduce variations within classes. Compared with traditional clustering methods, GCN can leverage the relationships between node neighbors to predict linkage likelihoods and obtain reliable clusters in unlabeled target domain. Second, although GCN performs well in clustering, it still fails to assign pseudo-labels to some images with extreme poses or expressions. In order to make better use of these remaining unlabeled samples and reduce intra-domain gap between pseudo-labeled and unlabeled samples, we propose an adversarial mutual information (MI) loss. It first moves the class prototypes, i.e., feature centers of classes, towards unlabeled target samples by minimizing MI with respect to target classifier, and then clusters all target samples around the updated prototypes by maximizing MI with respect to feature extractor. Through this min-max game, discriminative features can be obtained in the whole target domain.

Our contributions can be summarized into three aspects.

1) We introduce a new concept called intra-domain gap between pseudo-labeled and unlabeled target samples, and propose an adversarial MI loss to reduce this gap and enhance the discrimination ability of network for the whole target domain. Our loss can pull target prototypes towards the intermediate region between pseudo-labeled and unlabeled samples which makes unlabeled samples more easily clustered around the prototypes. We proved that our adversarial MI loss can compensate for the weakness of clustering based adaptation and further improve the model generalization.

2) To assist adversarial MI loss, we introduce GCN into UDA problem to mitigate variations within target classes. GCN is beneficial to the learning of neighborhood-invariance and could infer more accurate pseudo labels in unlabeled target domain by which the source model can be adapted.

3) Extensive experimental results on RFW \cite{wang2019racial}, IJB-A \cite{klare2015pushing} and IJB-C \cite{maze2018iarpa} datasets show that our AIN has the potential to generalize to a variety of scenarios and successfully transfers recognition knowledge cross image conditions and across faces with different poses and skin tones. It outperforms other UDA methods and offers a new state-of-the-art in performance on RFW \cite{wang2019racial} under UDA setting.

The remainder of this paper is structured as follows. In the next section, we review the related approaches of unsupervised domain adaptation and graph convolution network. In Section III, we propose the adversarial information network to improve cross-domain generalization for FR systems. In Section V, experimental results are shown and we validate the effectiveness of AIN method. Finally, we conclude and discuss the future work.

\section{Related work}




\subsection{Deep unsupervised domain adaptation}

Recently, many UDA approaches \cite{Long2016Unsupervised,courty2017optimal,8767033,8892738} are proposed to align the distributions of source and target domains. For example, DDC \cite{Tzeng2014Deep} and DAN \cite{Long2015Learning} used maximum mean discrepancies (MMD) to reduce the distribution mismatch. DANN \cite{Ganin2015Unsupervised} made the domain classifier fail to predict domain labels by a gradient reversal layer (GRL) such that feature distributions over the two domains are similar.
By borrowing the idea of self-training, other methods \cite{xie2018learning,chen2018progressive,8362753} generated pseudo-labels for target samples with the help of source model. For example,  MSTN \cite{xie2018learning} directly applied source classifier to generate target pseudo-labels and then aligned the centroids of source classes and target pseudo-classes. Zhang et al. \cite{zhang2018collaborative} selected reliable pseudo-samples with high classification confidence to finetune the source model, and simultaneously performed adversarial learning to learn domain-invariant features.

However, different from UDA in object classification, UDA in face recognition is a more complex and realistic setting where the source and target domains have completely disjoint classes/identities. Due to this challenge, very few studies have focused on it. Kan et al. \cite{kan2015bi,kan2014domain} proposed to shift source samples to target domain. Each shifted sample should be sparsely reconstructed by neighbors from target domain in order to force them to follow similar distributions. 
Sohn et al. \cite{sohn2017unsupervised} synthesized video frames, and utilized feature distillation and adversarial learning to align the image and video domains. MFR \cite{guo2020learning} synthesized the source/target domain shift with a meta-optimization objective, and applied hard-pair attention loss, soft-classification loss and domain alignment loss to learn domain-invariant features.
IMAN \cite{wang2019racial} applied a spectral clustering algorithm to generate pseudo-labels, and maximized MI between unlabeled target data and classifier's prediction to learn discriminative target features.

\subsection{Graph convolution network}

Recently, many works concentrate on GCN \cite{bruna2013spectral,kipf2016semi} for  handling graph data. Mimicking CNNs, modern GCNs learn the common local and global structural patterns of graphs through the designed convolution and aggregation functions. GraphSAGE \cite{hamilton2017inductive} made the GCN model scalable for huge graph by sampling the neighbors rather than using all of them. GAT \cite{velivckovic2017graph} introduced the attention mechanism to GCN to make it possible to learn the weight for each neighbor automatically.

In computer vision, GCNs have been successfully applied to various tasks and have led to considerable performance improvement. Xu et al. \cite{9397307} proposed a novel multi-stream attention-aware GCN for video salient object detection. Global-Local GCN \cite{zhang2020global} is proposed to perform global-to-local discrimination to select useful data in a noisy environment. Chen et al. \cite{chen2021learning} proposed a general GCNs based framework
to explicitly model inter-class dependencies for multi-label classification. Wang et al. \cite{wang2019linkage} proposed to construct instance pivot sub-graphs (IPS) that depict the context of given nodes and used the GCN to reason the linkage likelihood between nodes and their neighbors. Yang et al. \cite{yang2020learning} transformed the clustering problem into two sub-problems and designed GCN-V and GCN-E to estimate the confidence of vertices and the connectivity of edges.

Several works \cite{ma2019gcan,zhong2020learning,bai2021unsupervised} also introduced GCN into UDA and exploited the effect of GCN on reducing domain shifts. GCAN \cite{ma2019gcan} leveraged GCN to capture data structure information and utilized data structure, domain label, and class label jointly for adaptation.
GPP \cite{zhong2020learning} found reliable neighbors from a target memory by GCN and pushed these neighbors to be close to each other by a soft-label loss. MDIF \cite{bai2021unsupervised} proposed a domain-agent-node as the global domain representation and fused domain information to reduce domain distances by receiving information from other domains in GCN. In our paper, we directly utilize GCN to generate reliable pseudo-labels and tune the network by these labels. Benefiting from it, our adversarial MI loss can be performed stably and accurately such that intra-domain gap is reduced when adapting.

\section{Adversarial information network}

\begin{figure*}[htbp]
\centering
\includegraphics[width=17cm]{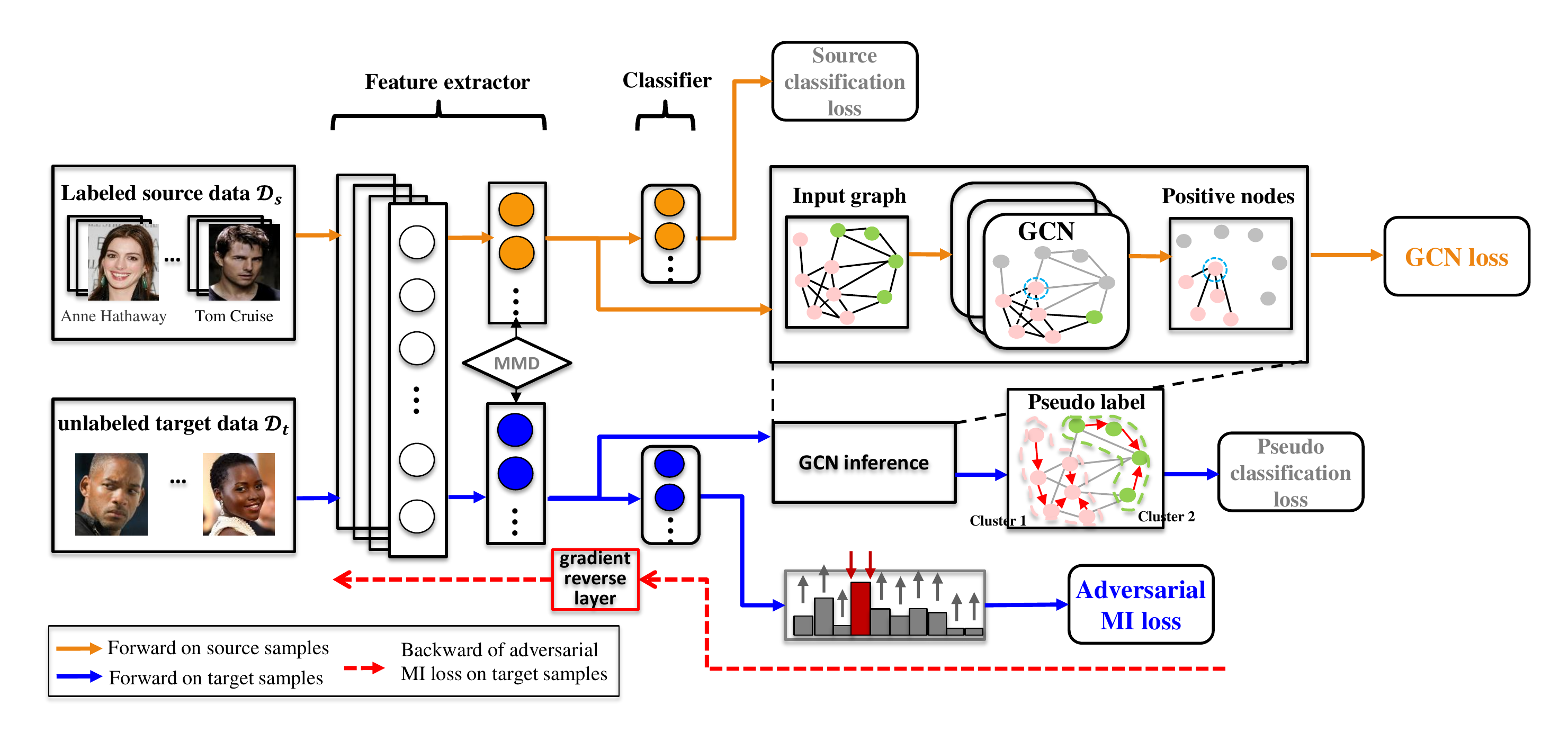}
\caption{Overview of AIN architecture. \textbf{GCN-based clustering.}  A GCN is trained on source domain to predict the positive and negative neighbors of the input. Then we use the learned GCN to infer linkage likelihoods between target data and generate pseudo labels.  These labels are utilized to optimize the feature extractor $F$ and target classifier $C_T$ with supervision of pseudo classification loss. \textbf{Adversarial MI loss.} To reduce intra-domain gap between pseudo-labeled and unlabeled target samples, $C_T$ is trained to minimize MI to move the prototypes towards unlabeled target samples whereas $F$ is trained to maximize MI to learn discriminative representations in the target feature space. }
\label{fig9} 
\end{figure*}

In our study, we denote the labeled source domain as $\mathcal{D}_{s}=\{{x^{s}_{i}},{y^{s}_{i}}\}^{M}_{i=1}$ where $x^{s}_{i}$ is the $i$-th source sample, $y^{s}_{i}$ is its category label, and $M$ is the number of source images. Similarly, the unlabeled target domain data are represented as $\mathcal{D}_{t}=\{{x^{t}_{i}}\}^{N}_{i=1}$ where $x^{t}_{i}$ is the $i$-th target sample and $N$ is the number of target images. The identities in source and target domains are completely different, and the identity annotations of target domain are not available. There is a discrepancy between the distributions of source and target domains $P(X_s,Y_s)\neq P(X_t,Y_t)$. Our goal is to enhance the discriminability of target domain through training on labeled $\mathcal{D}_{s}$ and unlabeled $\mathcal{D}_{t}$. 

\subsection{Overview of framework}

The architecture of our proposed AIN is shown in Fig. \ref{fig9}, which consists of a feature extractor $F$ and two classifiers $C_S$ and $C_T$. $C_S$ consists of weight vectors $W^s = \{w_1^s , w_2^s,...,w_{M_C}^s\}$ where $M_C$ represents the number of source classes. $C_T$ consists of weight vectors $W^t = \{w_1^t , w_2^t,...,w_{N_C}^t\}$ where $N_C$ represents the number of target pseudo-classes. After training, the directions of weight vectors should be representative to the features of the corresponding classes. In this respect, the learned weight vector can be regarded as prototype for each class.

The inputs are sampled from both source domain and target domain, and are fed into the network. 
First, we train the source classifier $C_S$ and the feature extractor $F$ with Softmax or Arcface \cite{deng2018arcface} loss to learn basic representations on source data. 
After pre-training, a GCN is trained with source features, and is inferred on target data to cluster images into pseudo classes. Then, we adapt feature extractor $F$ and target classifier $C_T$ with these generated pseudo-labels such that variations within target domain can be mitigated. However, supervision with pseudo-labeled target samples will make them clustered to their prototypes but separated from the remaining unlabeled samples, which results in intra-domain gap. To reduce this gap between pseudo-labeled and unlabeled target samples, we finally perform our adversarial MI learning in which $F$ and $C_T$ are learned in an adversarial min-max manner iteratively.

\subsection{GCN-based clustering}

Some latent variations in target domain are hard to explicitly discern without fine-grained labels. 
Inspired by \cite{zhong2020learning,zhang2020global,wang2019linkage}, we introduce GCN into adaptation problem of FR and generate pseudo-labels for adaptation training. 
The details are as follows.

\textbf{Input graph construction.} Given a source image $x^s_{i}$, we aim to construct a graph $\mathcal{G}^s_i(V^s_i,E^s_i)$ as input to train GCN, in which $V^s_i$ denotes the set of nodes and $E^s_i$ indicates the set of edges. 1) \textbf{Node discovery}. First, we feed this source data into pre-trained network and extract its deep feature ${\rm f}^s_i$. Then, we use its one-hop and two-hop neighbors as nodes for $\mathcal{G}^s_i(V^s_i,E^s_i)$, which is denoted as $V^s_i= \{v_1, v_2, ..., v_k \}$. Based on cosine similarity, one-hop neighbors can be found by selecting $k_1$ nearest neighbors of $x^s_{i}$ from other source data, and two-hop neighbors can be found by selecting $k_2$ nearest neighbors of one-hop neighbors. Note that $x^s_{i}$ itself is excluded from $V^s_i$. 2) \textbf{Node features}. We denote $\mathcal{F}^s_i=\{{\rm f}^s_{v_1},{\rm f}^s_{v_2}, ..., {\rm f}^s_{v_k}\}$ as the node features. In order to encode the information of $x^s_{i}$, we normalize $\mathcal{F}^s_i$ by subtracting ${\rm f}^s_i$,
\begin{equation}
\mathcal{F}^s_i=\{{\rm f}^s_{v_1}-{\rm f}^s_i,{\rm f}^s_{v_2}-{\rm f}^s_i, ..., {\rm f}^s_{v_k}-{\rm f}^s_i\},
\end{equation}
where $\mathcal{F}^s_i \in \mathbb{R}^{k\times d}$, and $d$ is the feature dimension. 3) \textbf{Edge linkage}. For each node $v \in V^s_i$, we first find its $k_3$ nearest neighbors from all source data, and then add edges between $v$ and its neighbors into the edge set $E^s_i$ if its neighbors appear in $V^s_i$. We denote these linked neighbors of node $v$ as $\mathcal{N}(v)$. As well, an adjacency matrix $\mathcal{A}^s_i \in \mathbb{R}^{k\times k}$ is computed to represent the weights of $E^s_i$,
\begin{equation}
\left ( \mathcal{A}^s_i\right )_{p,q}=\left ( \mathcal{F}^s_i\right )^T_p \left ( \mathcal{F}^s_i\right )_q, \forall p,q \in V^s_i.
\end{equation}
Finally, along with the adjacency matrix $ \mathcal{A}^s_i$ and node features $\mathcal{F}^s_i$, the topological structure of $\mathcal{G}^s_i(V^s_i,E^s_i)$ is constructed.

\textbf{GCN training.} Taken $ \mathcal{A}^s_i$, $\mathcal{F}^s_i$ and $\mathcal{G}^s_i(V^s_i,E^s_i)$ as input, we aim to leverage the context contained in graphs to predict if a node $v$ is positive (belongs to the same class with $x^s_{i}$) or negative (belongs to the different class with $x^s_{i}$). We apply GCN to achieve this goal. Every graph convolutional layer in the GCN can be written as a non-linear function and is factorized into feature aggregation and feature transformation.

First, feature aggregation updates the representation of each node $v \in V^s_i$ by aggregating the representations of its neighbors $\mathcal{N}(v)$ and then concatenating $v$'s representation with the aggregated neighborhood vector,
\begin{equation}
\hat{h}^{l}_{\mathcal{N}(v)}=\left [ h^{l}_v||g\left (h^{l}_{u \in \mathcal{N}(v)}\right ) \right ],
\end{equation}
where $h^{l}_v$ means the node $v$'s feature outputted by $l$-th GCN layer, and $h^{0}_v={\rm f}^s_{v}$. $||$ is the concatenation operator, and $g(\cdot ): \mathbb{R}^{\left | \mathcal{N}(v) \right |\times d_{in}}\rightarrow \mathbb{R}^{d_{in}}$ is a learnable aggregation function,
\begin{equation}
g\left (h^{l}_{u \in \mathcal{N}(v)}\right ) =\sum_{u \in \mathcal{N}(v)}\sigma \left ( W^l_{g} \left (s_{u,v}\cdot  h^{l}_u \right )+b^l_g \right ),
\end{equation}
where $\sigma$ denotes the activation function, and $W^l_{g} \in \mathbb{R}^{d_{in}\times d_{in}} $ and $b^l_{g}\in \mathbb{R}^{d_{in}\times 1}$ are learnable weight and bias matrixs in the $l$-th layer for aggregation. $s_{u,v}=\Lambda _{u}^{-\frac{1}{2}}\left ( \mathcal{A}^s_i\right )_{u,v}\Lambda _{v}^{-\frac{1}{2}}$ is the normalized cosine similarity between node $v$ and its neighbor $u$, and $\Lambda _{u}$ is the degree of node $u$ \cite{zhang2020global}.

Then, feature transformation transforms the aggregated representations via a fully connected layer with activation function $\sigma$,
\begin{equation}
h_v^{l+1}=\sigma \left ( W^l\cdot \hat{h}^{l}_{\mathcal{N}(v)} \right ),
\end{equation}
where $W^l \in \mathbb{R}^{2d_{in}\times d_{out}} $ is a learnable weight matrix in the $l$-th layer for transformation.

At the last layer, we utilize cross-entropy loss after the softmax activation to optimize GCN to predict the probability that a node $v$ belongs to the same class of $x^s_i$,
\begin{equation}
\mathcal{L}_{GCN}=-\frac{1}{\left | V^s_i \right |}\sum_{v \in V^s_i}\left ( y_v^s{\rm log}p_v^s+\left ( 1-y_v^s \right ){\rm log}\left ( 1-p_v^s \right ) \right ), \label{gcn}
\end{equation}
where $p_v^s$ is the predicted possibility and $y_v^s$ is the ground-truth label of node $v$.

\textbf{Inference on target data.} We infer the trained GCN on target domain, and obtain the predicted possibilities which indicate whether two target data belong to the same class. According to these possibilities, we can construct a large graph $\mathcal{G}^t(V^t,E^t)$ on all target data, where $V^t$ represents a set of all target images and $E^t$ denotes a set of edges weighted by the possibilities given by GCN. We cut the edges below the threshold $\eta $ and save the connected components as clusters (pseudo identities). In our method, $\eta $ varies from 0.1 to 1 in steps of 0.1. In the $0$-th iteration, we cut edges below $\eta =0.1$ and maintain connected clusters whose sizes are larger than a pre-defined maximum size in a queue to be processed in the next iteration. In the $1$-th iteration, $\eta $ for cutting edges is increased to 0.2. This process is iterated until the queue is empty or $\eta $ becomes 1. \emph{It is worth noting that the images in singleton clusters, i.e., clusters that contain only a single sample, are not assigned pseudo-labels by our GCN-based clustering, and are treated as unlabeled data in adversarial MI loss}. It is because they are probably noisy samples or they may cause long-tail problem during training. After that, we adapt the network on pseudo-labeled data with Softmax or Arcface \cite{deng2018arcface} such that variations within target domain can be reduced.

\subsection{Adversarial mutual information loss} \label{analysis of intra domain gap}

\begin{figure}
\centering
\subfigure[Softmax]{
\label{fig3a} 
\includegraphics[width=4cm]{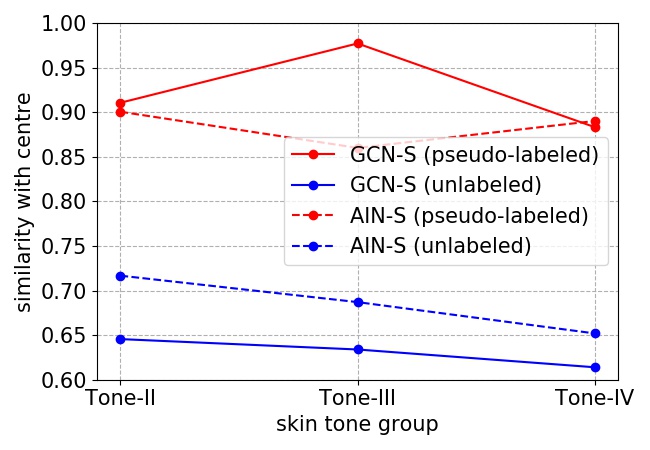}}
\hspace{0cm}
\subfigure[Arcface]{
\label{fig3b} 
\includegraphics[width=4cm]{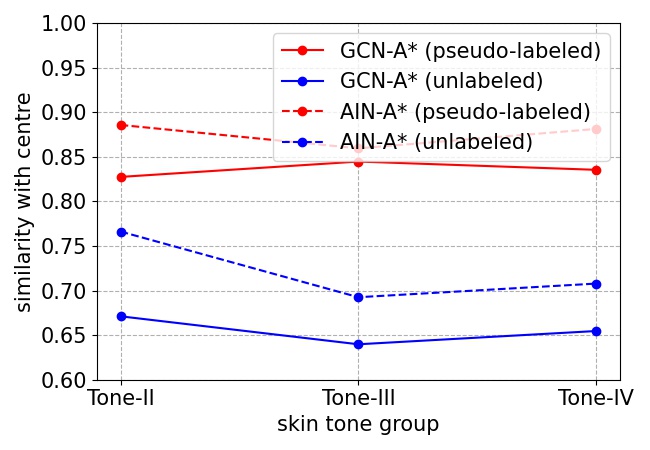}}
\caption{Intra-class distances of pseudo-labeled samples and unlabeled samples in target domains with Tone II-IV (larger is better). We can find that pseudo-labeled samples are always much closer to prototypes compared with unlabeled samples leading to gaps between them. }
\label{intra_domain_gap} 
\end{figure}

\textbf{Analysis of intra-domain gap.} In this paper, we rethink the clustering based adaptation method by investigating intra-domain gap between pseudo-labeled data and the remaining unlabeled samples within target domain. We find that, although GCN-based clustering can improve target performance preliminary, it discards some useful information in target domain and may cause intra-domain gap due to the imperfection of pseudo-labels. To investigate this, we extract target features by GCN-based clustering and compute intra-class distances for pseudo-labeled samples and unlabeled samples respectively, as shown in Fig. \ref{intra_domain_gap}. Intra-class distance represents the cosine similarity between samples and their corresponding centres, which can be formulated as:
\begin{equation}
D_g=\frac{1}{N_g}\sum_{i=1}^{N_g}\frac{1}{\left | \mathcal{I}_i \right |}\sum_{x_{j}\in \mathcal{I}_i}cos(x_{j},c_{i}), \label{intra_domain_distance}
\end{equation}
where $g\in\{p,u\}$ represents pseudo-labeled or unlabeled set, $N_g$ is the number of identities in $g$ set, $\mathcal{I}_i$ is the set of all images belonged to $i$-th identity, and $c_{i}$ is the feature centre of $i$-th identity. Domain adaptation across faces with different skin tones is performed by using (a) Softmax and (b) Arcface \cite{deng2018arcface} as pseudo classification loss, respectively. From the results plotted by solid lines, we can find that pseudo-labeled samples are always much closer to prototypes compared with unlabeled samples leading to gaps between them. This is because supervising with pseudo labels pulls these pseudo-labeled samples towards their feature clusters; while the remaining unlabeled samples which less correlate with the given pseudo-labeled samples are distant from these clusters and thus are more difficult to be attracted to them. With this analysis, it is natural to raise the following questions: How can we take full advantage of unlabeled target samples who are not assigned pseudo-labels by GCN to further optimize network and learn more discriminative representations? How can we mitigate this intra-domain gap and pull unlabeled samples towards their corresponding prototypes in an unsupervised manner? To address these issues, we propose a novel adversarial mutual information learning.

\textbf{Max-step.} We assume that there exists a prototype for each target class, which can be represented by weight vector $W^t = \{w_1^t , w_2^t,...,w_{N_C}^t\}$ of target classifier $C_T$. Inspired by \cite{wang2019racial,salimans2016improved,barratt2018note,Yuan2012Information}, mutual information is introduced as a regularization term to cluster target samples, no matter whether they are successfully assigned pseudo-labels or not, around their prototypes. It  can be equivalently expressed as: $I(\mathbf{X};\mathbf{O})=H(\mathbf{O})-H(\mathbf{O}|\mathbf{X})$, which aims to maximize mutual information between unlabeled target data $\mathbf{X}$ and classifier's prediction $\mathbf{O}$ under an unsupervised setting.

Maximizing MI can be broken into: 1) minimizing the  conditional entropy $H(\mathbf{O}|\mathbf{X})$ and 2) maximizing the marginal entropy $H(\mathbf{O})$. The first term makes the predicted possibility of each sample $p(o|x)$ look ``sharper" and confident. This exploits entropy minimization principle \cite{grandvalet2005semi} which is utilized in some UDA methods \cite{Long2016Unsupervised,saito2019semi,mei2020instance} to favor the low-density separation between classes. 
Meanwhile, the second term makes samples assigned evenly across the target categories of dataset, thereby avoiding degenerate solutions, i.e., most samples are classified to the same class. The corresponding MI loss is as follows:
\begin{equation}
\mathcal{L}_{M}=-\gamma \mathbb{E}_{o\sim \mathcal{P}\left ( O \right )}\left [ \log p\left ( o \right ) \right ]+\mathbb{E}_{x\sim \mathcal{P}\left ( X \right )}\left [\sum_{o} p\left ( o|x \right )\log p\left ( o|x \right ) \right ] \label{MI},
\end{equation}
where $\gamma$ is the trade-off parameter. $p\left ( o|x \right )$ represents the predicted possibility of target sample $x$ outputted by classifier. $\mathcal{P}\left ( \mathbf{O} \right )$ is the distribution of target category. Note that we utilize $N_C$, which is the number of target pseudo-clusters generated by GCN-based clustering, to approximate the ground-truth number of target categories. $p(o)$ can be approximated by averaging $p(o|x)$ in the minibatch: $p(o)=\mathbb{E}_{x\sim \mathcal{P}\left ( X \right )}\left [p\left ( o|x \right ) \right ]=\tfrac{1}{n}\sum_{x} p\left ( o|x \right )$.

\begin{figure}
\centering
\includegraphics[width=9cm]{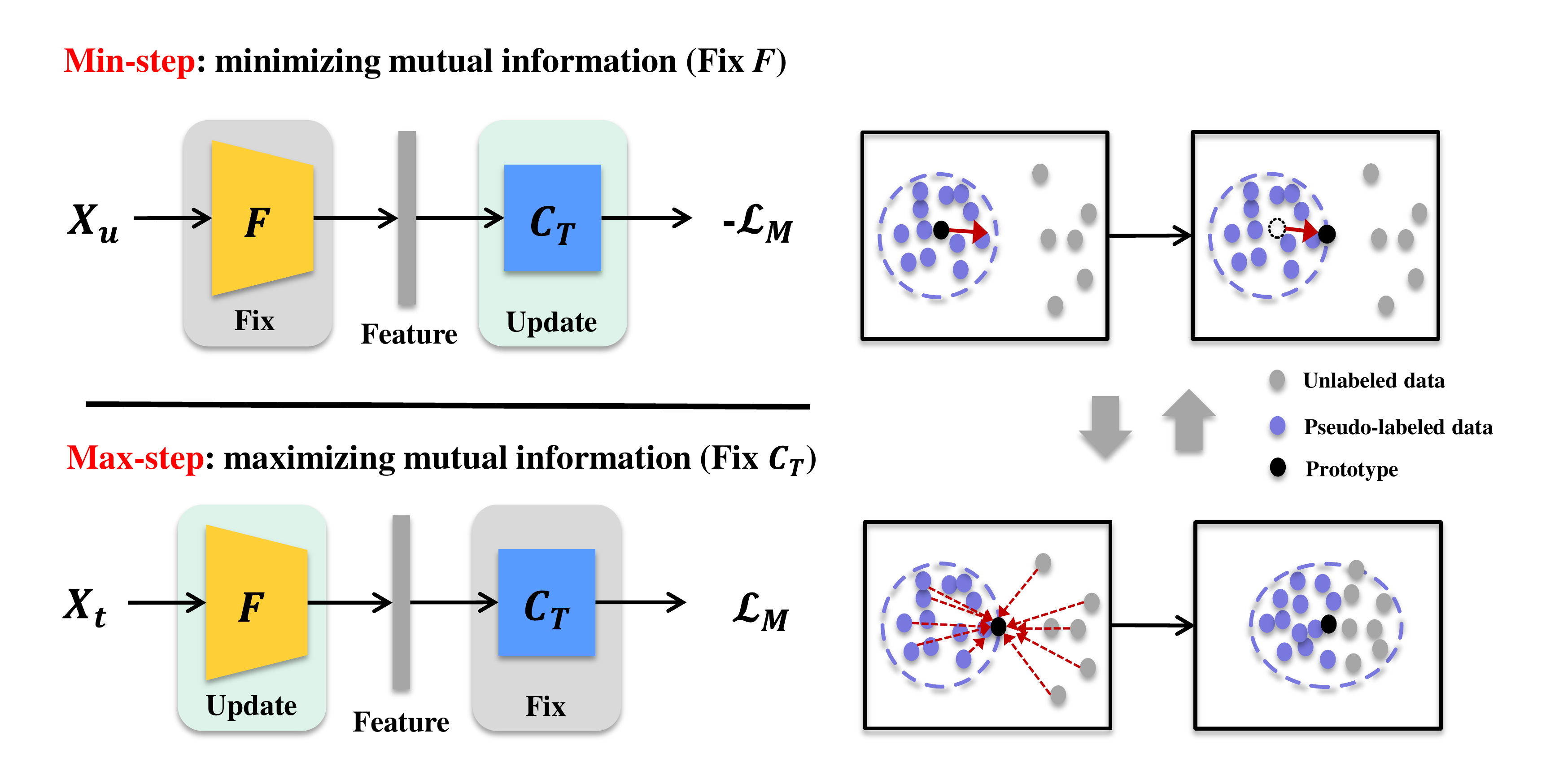}
\caption{ The proposed adversarial mutual information learning. }
\label{max_min}
\end{figure}

As we need to cluster target features around prototypes to obtain discriminative features on target domain, we optimize the feature extractor $F$ by maximizing mutual information to force the target features to be assigned to one of the prototypes,
\begin{equation}
\widehat{\theta }_F=arg \max\limits_{\theta_F}\ \lambda_2 \mathcal{L}_{M}(X_{t}), \label{max MI}
\end{equation}
where $\theta_F$ denotes the parameters of feature extractor.

\textbf{Min-step.} However, 
the presence of the intra-domain gap will prevent unlabeled samples from clustering towards prototypes. To reduce intra-domain gap, we propose to modulate the position of prototype by moving each $w_i^t$ towards unlabeled target features. As we know, after supervising with pseudo-labels, pseudo-labeled target samples and prototypes are related with each other and are located in high-MI region whereas unlabeled samples are discarded in low-MI region. Therefore, we can pull target prototypes towards the intermediate (middle-MI) region through minimizing MI with respect to target classifier since MI decreases as features move away from the prototypes while features far from the prototypes can be attracted towards prototypes. The objective function can be computed as follows,
\begin{equation}
\widehat{W^t}=arg \min\limits_{W^t}\ \lambda_1 \mathcal{L}_{M}(X_{u}) + \mathcal{L}_{T}(X_{p},\hat{Y}_{p}), \label{min MI}
\end{equation}
where $W^t$ is the weight vector of target classifier $C_T$. $X_{p}$ is the set of pseudo-labeled target samples and $\hat{Y}_{p}$ denotes the set of their corresponding pseudo-labels generated by GCN-based clustering. $X_{u}$ is the set of unlabeled target samples, $X_{p}\cap X_{u}=\Omega$ and $X_{p}\cup X_{u}=X_{t}$. $\mathcal{L}_{T}$ is pseudo-classification loss, i.e., Softmax or Arcface \cite{deng2018arcface}, computed on pseudo-labeled samples to prevent mutual information from decreasing too much so as to avoid the collapse of network. In this way, we move the class prototypes in a mutual information minimization direction such that unlabeled data can be clustered more easily.

To summarize, $F$ and $C_T$ are thus learned in an adversarial min-max manner iteratively and alternatively as shown in Fig. \ref{max_min}. First, we train $C_T$ by MI minimization to update prototypes. Then, after modulating the positions of prototypes, $F$ is trained by MI maximization to cluster target images to the updated prototypes, resulting in the desired discriminative features.

\begin{algorithm}[htb]
\caption{ Adversarial Information Network (AIN).}
\label{al1}
\begin{algorithmic}[1]
\REQUIRE ~~\\
Labeled source samples $\mathcal{D}_{s}=\{{x^{s}_{i}},{y^{s}_{i}}\}^{M}_{i=1}$, and unlabeled target samples $\mathcal{D}_{t}=\{{x^{t}_{i}}\}^{N}_{i=1}$.
\ENSURE ~~\\
Parameters of feature extractor $\theta_F$.
\STATE \textbf{\emph{Stage-1: // Pre-training:}}
\STATE  Initialize $\theta_F$ and $W^s$ on $\mathcal{D}_{s}$ and $\mathcal{D}_{t}$ by $\mathcal{L}_{S}(X_{s},Y_{s})$ and MMD \cite{Borgwardt2006Integrating,Long2015Learning};
\STATE \textbf{\emph{Stage-2: // GCN-based clustering:}}
\STATE Train GCN on $\mathcal{D}_{s}$ by $\mathcal{L}_{GCN}(X_{s},Y_{s})$;
\STATE Infer GCN on $\mathcal{D}_{t}$ to generate $\mathcal{D}_{p}=\{{x^{p}_{i}},{\hat{y}^{p}_{i}}\}^{P}_{i=1}$;
\STATE Adapt $\theta_F$, $W^s$ and $W^t$ on $\mathcal{D}_{s}$ and $\mathcal{D}_{p}$ by $\mathcal{L}_{S}(X_{s},Y_{s})$ and $\mathcal{L}_{T}(X_{p},\hat{Y}_{p})$;
\STATE \textbf{\emph{Stage-3: // Adversarial MI learning:}}
\STATE Adapt $\theta_F$ and $W^s$ on $\mathcal{D}_{s}$ and $\mathcal{D}_{t}$ by $\mathcal{L}_{S}(X_{s},Y_{s})$ and MMD, simultaneously, $\theta_F$ and $W^t$ are trained adversarialy by $\mathcal{L}_{A-MI}(X_{t},\hat{Y}_{p})$ according to Eq. \ref{min MI} and Eq. \ref{max MI} ;
\end{algorithmic}
\end{algorithm}

\subsection{Adaptation network}

The goal of training is to minimize the following loss:
\begin{equation}
\begin{split}
\mathcal{L}&=\mathcal{L}_{S}(X_{s},Y_{s})+\alpha\mathcal{L}_{T}(X_{p},\hat{Y}_{p})+  \mathcal{L}_{A-MI}(X_{t},\hat{Y}_{p})\\
&+\beta  \sum_{l\in\mathcal{L} }MMD^2(X_s^l,X_t^l) +\mathcal{L}_{GCN}(X_{s},Y_{s}), \label{eq2}
\end{split}
\end{equation}
where $\alpha$ and $\beta$ are the parameters for the trade-off between different terms. We adopt a three-stage training scheme including pre-training, GCN-based clustering and adversarial MI learning as shown in Algorithm \ref{al1}.

\textbf{Stage-1: pre-training.} We pre-train the network on source and target data by source classification loss $\mathcal{L}_{S}(X_{s},Y_{s})$ and MMD loss $MMD^2(X_s^l,X_t^l)$ \cite{Borgwardt2006Integrating,Long2015Learning}. MMD loss is a commonly-used global alignment method adopted on source and target features $X_*^l$ of $l$-th layer. Through minimizing MMD, inter-domain discrepancy can be reduced.

\textbf{Stage-2: GCN-based clustering.} Source features are extracted by the pre-trained model and input graphs are constructed. We train GCN on source data by GCN loss $\mathcal{L}_{GCN}(X_{s},Y_{s})$, and then infer the trained GCN model on target domain to generate pseudo labels $\hat{y}^p$. The pre-trained model is finetuned on source data and pseudo-labeled target data $\mathcal{D}_{p}=\{{x^{p}_{i}},{\hat{y}^{p}_{i}}\}^{P}_{i=1}$ by $\mathcal{L}_{S}(X_{s},Y_{s})$ and pseudo classification loss $\mathcal{L}_{T}(X_{p},\hat{Y}_{p})$.

\textbf{Stage-3: adversarial MI learning.} We adapt the network on all source and target data using $\mathcal{L}_{S}(X_{s},Y_{s})$, $MMD^2(X_s^l,X_t^l)$ and our proposed adversarial MI loss $\mathcal{L}_{A-MI}(X_{t},\hat{Y}_{p})$.

\subsection{Discussion}

\textbf{Difference with information bottleneck (IB) theory.} IB \cite{tishby2000information} aims to learn more informative features based on a tradeoff between concise representation and good predictive power. It can be formulated as $I(Z;Y)-\beta I(Z;X)$, which maximizes MI between the learned features $Z$ and the labels $Y$, and simultaneously minimizes MI between $Z$ and the inputs $X$. However, without any label information, our MI loss is unable to maximize $I(O;Y)$. To make correct predictions in an unsupervised manner, our MI loss aims to learn information from $X$ as much as possible and thereby learn discriminative features by maximizing $I(O;X)$.

\textbf{Relation to curriculum learning (CL).} CL \cite{bengio2009curriculum} progressively assigns pseudo-labels for reliable target samples and utilizes them to optimize networks \cite{chen2018progressive}. Although supervision with pseudo-labeled samples can be of benefit to unlabeled samples, there is no doubt that this will bring more gains for pseudo-labeled samples compared with unlabeled samples in each iteration. Moreover, due to the particularity of FR, harder samples with large variances (e.g., pose, expression or occlusion) are still much more difficult to learn compared with easier samples which also leads to intra-domain gap. As proved in Section \ref{visual experiment}, mitigating intra-domain gap in CL by using our adversarial MI learning can further improve its target performance in FR. It is also worth noting that the idea of CL is actually not applied in our method, that is, GCN-based clustering and adversarial MI learning are not trained iteratively. Therefore, the remained unlabeled data have no more chances to be assigned labels in later stages and are optimized by unsupervised loss. Our AIN should emphasize more on reducing the negative effect of intra-domain gap.

\section{Experiments}

In our paper, the proposed method is evaluated under two face recognition scenarios, i.e., domain adaptation across faces with different skin tones, and domain adaptation across poses and image conditions. The details are as follows.

\subsection{Datasets}

\textbf{BUPT-Transferface.} BUPT-Transferface \cite{wang2019racial} is a training dataset for transferring knowledge across races. Considering that the concept of race is a social construct without biological basis \cite{buolamwini2018gender,yudell2016taking}, we aim to study this issue on the basic of skin tone. As skin tone was also taken into consideration when BUPT-Transferface was constructed, we change the names of four subsets from ``Caucasian, Asian, Indian and African" to ``Tone I-IV" according to their skin tones. Skin gradually darkens with the increase of tone value (from I to IV). There are 500K labeled images of 10K identities in lighter-skinned (Tone-I) set, and 50K unlabeled images in each darker-skinned (Tone II-IV) set. In our paper, we use labeled images of Tone-I subjects as source data, and use unlabeled images of darker-skinned subjects as target data.

\textbf{RFW.} RFW \cite{wang2019racial} is a testing database which can evaluate the performances of different races. It contains four testing subsets corresponding with BUPT-Transferface, and we also rename them ``Tone I-IV". Each subset contains about 3000 individuals with 6000 image pairs for face verification. 
We use RFW dataset to validate the effectiveness of our AIN method on transferring knowledge across faces with different skin tones. 

\textbf{CASIA-Webface.} CASIA-WebFace dataset \cite{yi2014learning} contains 10,575 subjects and 494,414 images. It is a large scale face dataset collected from website and consists of images of celebrities on formal occasions. Therefore, the faces of images are high-definition, frontal, smiling and beautiful. 

\textbf{IJB-A/C.} IJB-A \cite{klare2015pushing} and JB-C \cite{maze2018iarpa} are different from CASIA-WebFace and contain large number of images with low-definition and large pose. IJB-A database contains 500 subjects with 5,397 images and 2,042 videos; and IJB-C dataset consists of 31.3K images and 11,779 videos of 3,531 subjects.
We use CASIA-WebFace \cite{yi2014learning} as source data and use IJB-A/C as target data to validate the effectiveness of our AIN on reducing domain shifts cased by pose and image quality.

\subsection{Experimental Settings}

We adopt the similar ResNet-34 architecture described in \cite{deng2018arcface} as the backbone of feature extractor. For preprocessing, we use five facial landmarks for similarity transformation, then crop and resize the faces to 112$\times$112. Each pixel ([0, 255]) in RGB images is normalized by subtracting 127.5 and then being divided by 128.

\textbf{Stage-1: pre-training.} We follow the settings in DAN \cite{Long2015Learning} to perform MMD. A Gaussian kernel $k(x_{i},x_{j})=e^{-\left \| x_{i}-x_{j} \right \|^{2}/\mu}$ with the bandwidth $\mu$ is used. The bandwidth is set to be $\mu_{m}\cdot(1, 2^1, 2^2, 2^3, 2^4)$, respectively, where $\mu_{m}$ is the median pairwise distances on the training data. And we apply MMD on the last two fully-connected layers. The learning rate starts from 0.1 and is divided by 10 at 80K, 120K, 155K iterations. We finish the training process at 180K iterations. The batch size, momentum, and weight decay are set to be 200, 0.9 and $5\times 10^{-4}$, respectively.

\textbf{Stage-2: GCN-based clustering.} A GCN contained five graph convolutional layers with 256-dimension hidden features is adopted. To train it on source data, 5-NN input graphs are constructed by setting $k_1=80$, $k_2=5$ and $k_3=5$ according to \cite{wang2019linkage}. The SGD learning rate, weight decay and graph batch size are $10^{-3}$, $5\times 10^{-4}$ and 50. We train GCN for 20K iterations. Then, the generated pseudo-labeled data are utilized to finetune the network. The learning rate, batch size, momentum and weight decay are $10^{-3}$, 200, 0.9 and $5\times 10^{-4}$, respectively. We finish the training process at 50K iterations.

\textbf{Stage-3: adversarial MI learning.} Following \cite{wang2019racial}, the trade-off parameter $\gamma$ in Eq. \ref{MI} is set as 0.2. The learning rate, batch size, momentum, and weight decay are $10^{-4}$, 200, 0.9 and $5\times 10^{-4}$, respectively. The training process is finished at 20K iterations.

To validate the effectiveness of our AIN method, we apply it based on Softmax and Arcface loss \cite{deng2018arcface}. In AIN-S, Softmax is used as both source classification loss and pseudo classification loss, and the parameter $\alpha$, $\beta$, $\lambda_1$ and $\lambda_2$ are set to be 0.2, 2, 0.1 and 5. In AIN-A, Arcface \cite{deng2018arcface} is used as source classification loss and Softmax is used as pseudo classification loss; in AIN*-A, Arcface is used as both source classification loss and pseudo classification loss. The parameter $\alpha$, $\beta$, $\lambda_1$ and $\lambda_2$ are set to be 1, 10, 0.5 and 25, respectively.


\subsection{Experimental result}

\textbf{Domain adaptation across skin tone.} Existing training datasets usually contain large number of lighter-skinned (Tone I) people, but the images of darker-skinned (Tone II-IV) subjects are rare. The model trained on lighter-skinned people cannot generalize well on darker-skinned subjects leading to serious racial bias \cite{wang2020mitigating,wang2021meta} in face recognition. Domain adaptation across faces with different skin tones attempts to adapt knowledge from lighter-skinned subjects to darker-skinned ones. 3 adaptation scenarios are adopted, i.e., I$\rightarrow $II, I$\rightarrow $III and I$\rightarrow $IV. We train the models using BUPT-Transferface, and evaluate them on RFW \cite{wang2019racial}.

\begin{table}[htbp]
	\begin{center}
    \caption{Verification accuracy (\%) on RFW dataset when Softmax is utilized as source classification loss. Skin gradually darkens with the increase of tone value (from I to IV).}
    \label{iman-s}
    \small
    \setlength{\tabcolsep}{3.5mm}{
	\begin{tabular}{l|ccc}
		\hline
         Methods & II & III & IV \\ \hline \hline
         Softmax & 84.60 & 88.33  & 83.47 \\
         DDC \cite{Tzeng2014Deep} & 86.32 &  90.53  & 84.95 \\
         DAN \cite{Long2015Learning} &85.53 & 89.98  & 84.10 \\
         CORAL \cite{sun2016deep} & 86.57 & 91.02 & 85.03 \\
         FGAN \cite{ge2020fgan} & 83.70 & 88.28 & 83.62 \\
         BSP \cite{chen2019transferability} & 87.25 & 90.73 & 86.25 \\
         GPP \cite{zhong2020learning} & 89.68 & 91.65 & 89.42 \\
         IMAN-S \cite{wang2019racial} & 89.88 & 91.08  & 89.13\\ \hline
         \textbf{AIN-S (ours)} & \textbf{89.95} &  \textbf{93.73}  & \textbf{90.43} \\ \hline\hline
	\end{tabular}}
    \end{center}
\end{table}

\begin{table}[htbp]
	\begin{center}
    \caption{Verification accuracy (\%) on RFW dataset when Arcface is utilized as source classification loss. Skin gradually darkens with the increase of tone value (from I to IV).}
    \label{iman-a}
    \small
    \setlength{\tabcolsep}{3.5mm}{
	\begin{tabular}{l|ccc}
		\hline
         Methods & II & III & IV \\ \hline \hline
         Arcface \cite{deng2018arcface} & 86.27 & 90.48  & 85.13  \\
         DDC \cite{Tzeng2014Deep}  &87.55 & 91.63  & 86.28 \\
         DAN \cite{Long2015Learning} & 87.78 &  91.78  & 86.30  \\
         CORAL \cite{sun2016deep} & 87.93 & 92.18 & 86.50 \\
         FGAN \cite{ge2020fgan} & 86.20 & 90.53 & 85.95 \\
         BSP \cite{chen2019transferability} & 86.90 & 91.58 & 85.15 \\
         GPP \cite{zhong2020learning} & 90.60 & 93.08 & 88.75 \\
         IMAN-A \cite{wang2019racial} & 89.87 &  93.55  & 88.88 \\
         \textbf{AIN-A (ours)} & \textbf{92.47} &  \textbf{93.93}  & \textbf{91.28} \\ \hline
         IMAN*-A \cite{wang2019racial} & 91.15 &  94.15  & 91.42 \\
         \textbf{AIN*-A (ours)}& \textbf{93.58} &  \textbf{95.30}  & \textbf{93.02} \\ \hline
	\end{tabular}}
    \end{center}
\end{table}

From the results shown in Table \ref{iman-s} and \ref{iman-a}, we have the following observations. First, when trained on Tone-I training set and tested on Tone-I testing set, Softmax and Arcface \cite{deng2018arcface} achieve high accuracies of 94.12\% and 94.78\%. However, we observe a serious drop in performance when they are directly applied on darker-skinned subjects. 
This decline in accuracy is mainly caused by domain shift between faces with different skin tones. Second, global alignment methods, e.g., DDC \cite{Tzeng2014Deep}, DAN \cite{Long2015Learning} and BSP \cite{chen2019transferability}, don't take variations into consideration and thus only obtain limited improvement in target domain. Third, IMAN \cite{wang2019racial} utilized spectral-clustering based adaptation and MI maximization to learn discriminative target representations; GPP \cite{zhong2020learning} found reliable neighbors by GCN and pushed these neighbors to be close to each other by a soft-label loss. They achieve superior results on RFW which demonstrates the importance of dealing with variations in target domain. Fourth, our AIN approach clearly exceeds other compared methods and even outperforms IMAN \cite{wang2019racial} and GPP \cite{zhong2020learning}. For example, our AIN-S achieves about 6\% gains over Softmax model, and our AIN-A method produces 92.47\% and 91.28\% when tested on Tone-II and Tone-IV subjects, which is higher than IMAN-A \cite{wang2019racial} by 2.6\% and 2.4\%, respectively. It is because our method additionally takes intra-domain gap into consideration while IMAN \cite{wang2019racial} and GPP \cite{zhong2020learning} not. Benefitting from GCN-based clustering and adversarial learning, our AIN further improves the performances on darker-skinned subjects.

\textbf{Domain adaptation across pose and image condition.} In real world applications of face recognition, many factors, e.g., pose, illumination and image quality, can also cause the mismatch of distribution between training and testing samples. To evaluate the effectiveness and robustness of our AIN method, we perform domain adaptation experiments across poses and image conditions. CASIA-Webface \cite{yi2014learning}, IJB-A \cite{klare2015pushing} and IJB-C \cite{maze2018iarpa} datasets are employed to simulate this scenario: using CASIA-Webface with high-definition and frontal faces as source domain and using IJB-A/C with low-definition and large-pose faces as target domain. We use labeled CASIA-Webface and unlabeled IJB-A to train our AIN, and evaluate the trained models on IJB-A and IJB-C. 

\begin{table}
\begin{center}
\caption{Verification performance (\%) of IJB-A. ``Verif" represents the 1:1 verification and ``Identif." denotes 1:N identification.}
\label{ijba}
\small
\setlength{\tabcolsep}{1.5mm}{
\begin{tabular}{l|ccc|cc}
\hline
\multirow{3}{*}{Method} & \multicolumn{3}{c|}{IJB-A: Verif.}&\multicolumn{2}{c}{\multirow{2}{*}{IJB-A: Identif.}}\\
&\multicolumn{3}{c|}{TAR@FAR's of}\\ \cline{2-6}
& 0.001 & 0.01 & 0.1 & Rank1 & Rank10\\ \hline\hline
Bilinear-CNN \cite{chowdhury2016one} & - & - & - & 58.80 & - \\
Face-Search \cite{wang2015face} & - & 73.30 & -  & 82.00 & - \\
Deep-Multipose \cite{abdalmageed2016face} & - & 78.70 & -  & 84.60 & 94.70 \\
Triplet-Similarity \cite{sankaranarayanan2016triplet} & - & 79.00 & 94.50 & 88.01 & 97.38 \\
Joint Bayesian \cite{chen2016unconstrained} & - & 83.80 & - & 90.30 & 97.70 \\
Arcface \cite{deng2018arcface} &74.19 & 87.11& 94.87& 90.68&  96.07    \\
DAN-A \cite{Long2015Learning} &80.64  & 90.87& 96.22& 92.78& 97.01 \\
IMAN-A \cite{wang2019racial} & \textbf{84.19} & 91.88 & 97.05 & 94.05 & 98.04  \\  \hline
\textbf{AIN-A (ours)}& 83.04 & \textbf{92.51} & \textbf{97.48} & \textbf{94.60} & \textbf{98.22}  \\ \hline
\end{tabular}}
\end{center}
\end{table}

\begin{table}
\small
\begin{center}
\caption{Performance evaluation on the IJB-C dataset.}
\label{IJBC}
\setlength{\tabcolsep}{1.5mm}{
\begin{tabular}{l|ccc|cc}
\hline
\multirow{3}{*}{\textbf{Method}} & \multicolumn{5}{c}{\textbf{IJB-C}} \\ \cline{2-6}
&\multicolumn{3}{c|}{\textbf{ Verification TAR@FAR}} & \multicolumn{2}{c}{\textbf{ Identification}}\\
& 0.001 & 0.01 & 0.1 & Rank-1 & Rank-10  \\\hline
GOTS-2 \cite{maze2018iarpa}& 32.00 & 62.00 & 80.00 & -   & - \\
FaceNet \cite{schroff2015facenet} & 66.00 & 82.00 & 92.00 & - & - \\
DR-GAN \cite{tran2017disentangled}  & 66.10 & 82.40 & - & 70.80 &  82.80 \\
VGG \cite{parkhi2015deep} & 75.00  &  86.00  &  95.00  &   -  &  - \\
Yin et al. \cite{yin2019towards} & 75.60 & 89.20 & - & 77.60  & 86.10 \\
Arcface\tnote{1} \cite{deng2018arcface}& 88.88  &  94.76  &  98.10  &  88.05    & 93.56  \\ \hline
\textbf{AIN-A (ours)} & \textbf{88.93} & \textbf{95.08} & \textbf{98.32} & \textbf{88.15} & \textbf{93.64} \\
\hline
\end{tabular}}
\end{center}
\end{table}

The evaluation results are shown in Table \ref{ijba} and \ref{IJBC}. The superiority of our method can be also observed when using IJB-A and IJB-C as target domains. Arcface \cite{deng2018arcface}, which reported SOTA performances on the LFW \cite{huang2008labeled}, MegaFace \cite{kemelmacher2016megaface} and IJB-A/C challenges, also suffers from domain gap, while our adaptation method successfully outperforms Arcface and IMAN \cite{wang2019racial}. For example, after adaptation, our method achieves rank-1 accuracy = 94.6\% and TAR = 83.04\% at FAR of 0.001 when tested on IJB-A, which is higher than Arcface \cite{deng2018arcface} by 3.92\% in rank-1 accuracy and by 8.85\% in TAR at FAR of 0.001. The results further demonstrate the advantage of our AIN method.

\subsection{Ablation study}

\textbf{Effectiveness of each component.} 
To evaluate the effectiveness of each component, we train ablation models by removing some of them. 
The results of ablation study are shown in Table \ref{tab6}. From the results, we can find that each component makes an important contribution to AIN in terms of model generalization.

\begin{table}
	\begin{center}
    \caption{Ablation study on RFW dataset. GCN denotes our GCN-based clustering, and A-MI denotes our adversarial mutual information learning.}
     \label{tab6}
    \small
    \setlength{\tabcolsep}{1.5mm}{
	\begin{tabular}{c|ccc|ccc}
		\hline
         Methods  & MMD & GCN & A-MI & II & III & IV \\ \hline \hline
         Softmax  & & & & 84.60 &88.33  & 83.47  \\\hline
         \multirow{3}{*}{AIN-S}  & \checkmark & & &85.53 &89.98  & 84.10  \\
         &\checkmark &\checkmark & & 89.93 &92.75  & 90.07 \\
         &\checkmark &\checkmark &\checkmark & \textbf{89.95} & \textbf{93.73}  & \textbf{90.43}  \\ \hline\hline
         Arcface \cite{deng2018arcface}  & & & & 86.27 &90.48  & 85.13  \\\hline
         \multirow{3}{*}{AIN-A}  & \checkmark & & & 87.78 &91.78  & 86.30  \\
         &\checkmark &\checkmark & & 89.95 &92.98  & 89.03 \\
         &\checkmark &\checkmark &\checkmark & \textbf{92.47} & \textbf{93.93}  & \textbf{91.28}  \\ \hline
         \multirow{3}{*}{AIN*-A}  & \checkmark & & & 87.78 &91.78  & 86.30  \\
         &\checkmark &\checkmark & & 92.13 &94.38  & 91.86 \\
         &\checkmark &\checkmark &\checkmark & \textbf{93.58} & \textbf{95.30}  & \textbf{93.02}  \\ \hline
	\end{tabular}}
    \end{center}
\end{table}

1) GCN-based clustering. 
When adding GCN-based clustering, our AIN further improves target performance compared to MMD, which proves its effectiveness. Moreover, GCN-based clustering is a necessary preprocessing for adversarial MI learning. 
On the one hand, the number of target pseudo-categories is needed in MI loss. On the other hand, MI loss places more trust in the model prediction to move target data towards their nearest prototypes and make predictions look ``sharper". GCN-based clustering can initialize target classifier and guarantee its accuracy for MI.

2) Adversarial MI learning. It is observed that the recognition performance drops if our adversarial MI (A-MI) learning is discarded, which proves that our adversarial MI loss can compensate for the weakness of GCN-based clustering. For example, the performance of Tone-II subjects decreases from 92.47\% to 89.95\% when A-MI is omitted in AIN-A. This is because: (1) Adversarial MI learning can utilize the missing information of GCN-based clustering. (2) Adversarial MI learning can pull all target data towards their prototypes and learn more discriminative features regardless of intra-domain gap.  

\begin{figure}
\centering
\subfigure[Softmax]{
\label{fig3a} 
\includegraphics[width=4cm]{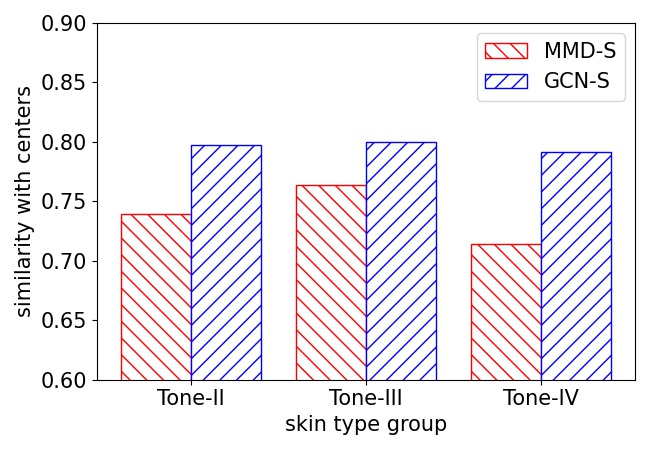}}
\hspace{0cm}
\subfigure[Arcface]{
\label{fig3b} 
\includegraphics[width=4cm]{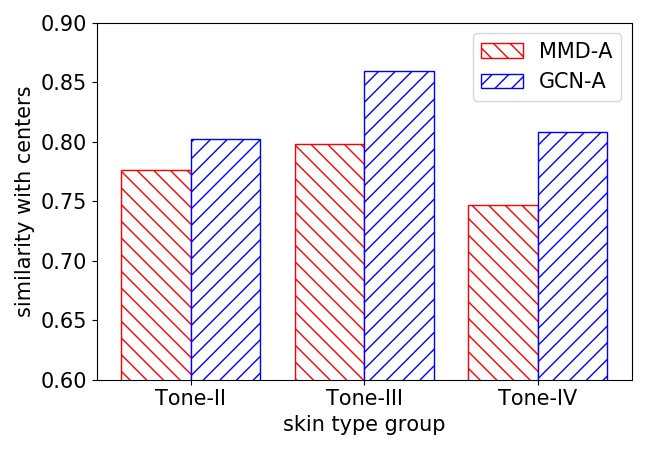}}
\caption{Intra-class compactness within target domain. We compute cosine similarities between all target samples and their corresponding prototypes for each target domain (larger is better).}
\label{local_variations} 
\end{figure}

\textbf{Effectiveness of mitigating variances.} We compute intra-class compactness of target features learned by MMD method and GCN-based clustering respectively, and compare them in Fig. \ref{local_variations}. Intra-class compactness refers to cosine similarity between all target samples and their corresponding centres in each target domain. As seen from the results, we proves that GCN-based clustering is an effective way to reduce variations within each class and lean more compact clusters. Unfortunately, it focuses more on the pseudo-labeled samples and less on the unlabeled ones in finetuning process and thus causes intra-domain gap in feature space as we analyze in Section \ref{analysis of intra domain gap}.

\begin{figure}
\centering
\subfigure[Softmax]{
\label{fig3a} 
\includegraphics[width=4cm]{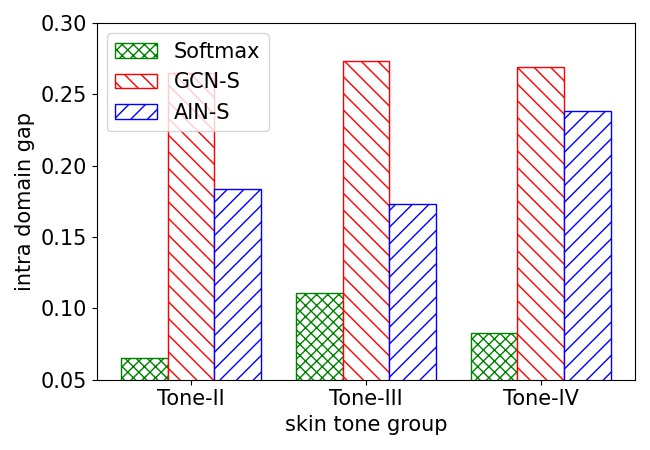}}
\hspace{0cm}
\subfigure[Arcface]{
\label{fig3b} 
\includegraphics[width=4cm]{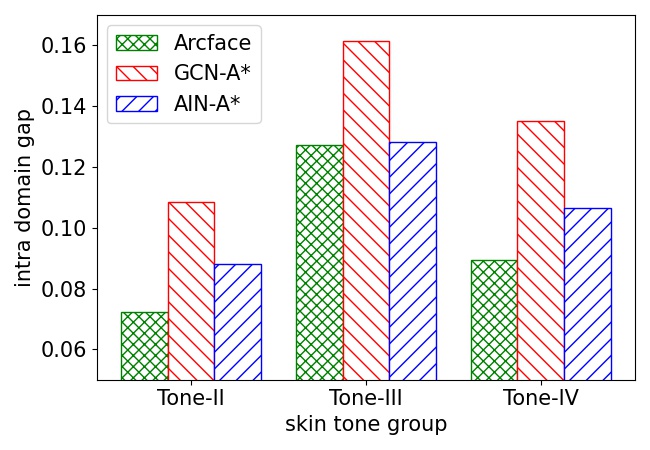}}
\caption{Intra-domain gap between pseudo-labeled and unlabeled samples within target domain (lower is better).}
\label{intra_domain_gap_reduce} 
\end{figure}

\textbf{Effectiveness of mitigating intra-domain gap.} The difference of intra-class compactness between pseudo-labeled samples and unlabeled samples is utilized as a criterion to evaluate intra-domain gap in our paper. It can be formulated as: $G=D_p-D_u$, where $D_p$ and $D_u$ are intra-class distances of pseudo-labeled data and unlabeled data respectively, and can be computed by Eq. \ref{intra_domain_distance}. A larger $G$ means that pseudo-labeled samples are much closer to prototypes leading to larger intra-domain gap. Two observations  can be obtained as follows. First, as shown in Fig. \ref{intra_domain_gap}, our adversarial MI learning can further improve both $D_p$ and $D_u$ compared with GCN-based clustering and learn more discriminative features. Second, from the results in Fig. \ref{intra_domain_gap_reduce}, we can see that GCN-based clustering model indeed has larger intra-domain gap than pre-trained model does, and that our adversarial MI learning effectively mitigates this gap. For example, our AIN-S reduces $G$ from 0.27 to 0.17 on Tone-III set compared with GCN-S. By pulling target prototypes towards the intermediate (middle-MI) region, our AIN can attract unlabeled samples to prototypes more easily and thus significantly improves their intra-class compactness.

\begin{table*}
\caption{Comparison with other clustering methods in terms of BCubed Precision, Recall, F-measure and the ratio of pseudo-labeled images to the whole data.}
\label{f-measure}
\small
	\begin{center}
	\begin{tabular}{l|cccc|cccc|cccc}
		\hline
         \multirow{2}{*}{Method}  &  \multicolumn{4}{c|}{Tone-II} & \multicolumn{4}{c|}{Tone-III} & \multicolumn{4}{c}{Tone-IV} \\ \cline{2-13}
          &  Precision & Recall & F-score & Ratio & Precision & Recall & F-score & Ratio & Precision & Recall  & F-score & Ratio  \\ \hline \hline
         K-means \cite{lloyd1982least} & 84.26 & 74.23 & 78.93 & 100.00 & 90.34 & 88.39 & 89.35 & 100.00 & 82.28 & 79.60 & 80.92 & 100.00 \\
         DBSCAN \cite{ester1996density} & 96.68 & 98.30 & 97.69 & 67.63 & 94.74 & 98.57 & 96.62 & 84.49 & 98.99 & 93.18 & 95.99 & 66.60 \\
         Spectral \cite{wang2019racial} & 99.08 & 90.20 & 94.43 & 51.93 & 99.94 & 91.62 & 95.60 & 53.26 & 96.91 & 91.09 & 93.91 & 52.95 \\
         \textbf{GCN-A (ours)} & 98.29 & 99.15& 98.72 & 71.69 & 99.14 & 99.57 & 99.35 & 79.27  & 98.28& 99.09 & 98.68 & 67.10 \\ \hline
	\end{tabular}
    \end{center}
\end{table*}

\textbf{Comparison with other clustering methods.} We compare our GCN method with other clustering algorithms in Table \ref{f-measure}. 
BCubed Precision, Recall, F-Measure \cite{amigo2009comparison} are reported. Considering that the images in singleton clusters are filtered and treated as unlabeled data by our AIN method, we also report the ratio of pseudo-labeled images to the whole data. As we can see from Table \ref{f-measure}, our method improves the precision and recall simultaneously. Compared with spectral clustering method in IMAN \cite{wang2019racial}, our GCN improves F-measure from 94.43\% to 98.72\% on Tone-II set, and from 93.91\% to 98.68\% on Tone-IV set, and it always generates less singleton clusters. 
Furthermore, we adapt the pre-trained model by using the pseudo-labels generated by spectral clustering method \cite{wang2019racial} and our GCN, and compare their accuracies on RFW \cite{wang2019racial} in Table \ref{gcn-da}. As we can see, benefitting from high-quality pseudo labels, our clustering method can obtain better adaptation performances on all testing sets, which proves the effectiveness of our clustering method.

\begin{table}[h]
	\begin{center}
    \caption{Verification accuracy (\%) of models adapted by pseudo-labels which are generated by different methods.} 
     \label{gcn-da}
    \small
	\begin{tabular}{l|cccc}
		\hline
         Methods &  II & III & IV \\ \hline \hline
         Softmax & 84.60 & 88.33  & 83.47 \\\hline
         Spectral-S \cite{wang2019racial} & 89.02 & 90.58  & 89.13 \\
         \textbf{GCN-S (ours)} & \textbf{89.93} & \textbf{92.75}  & \textbf{90.07} \\ \hline\hline
         Arcface \cite{deng2018arcface}  & 86.27 &90.48  & 85.13 \\ \hline
         Spectral-A \cite{wang2019racial} & 88.80 & 92.08  & 88.12 \\
         \textbf{GCN-A (ours)} & \textbf{89.95} & \textbf{92.98}  & \textbf{89.03} \\ \hline
         Spectral*-A \cite{wang2019racial} & 90.35 & 93.32  & 90.60\\
         \textbf{GCN*-A (ours)} & \textbf{92.13} & \textbf{94.38}  & \textbf{91.86} \\ \hline
	\end{tabular}
    \end{center}
\end{table}

\begin{figure*}
\centering
\subfigure[Arcface]{
\label{fig3a} 
\includegraphics[width=4.3cm]{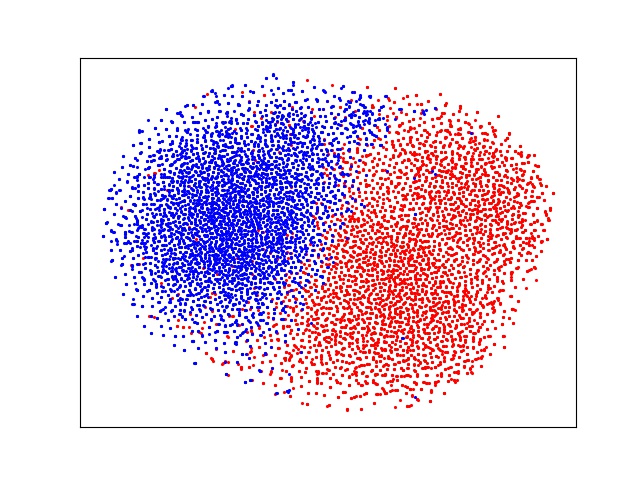}}
\hspace{0cm}
\subfigure[GCN-based clustering]{
\label{fig3b} 
\includegraphics[width=4.3cm]{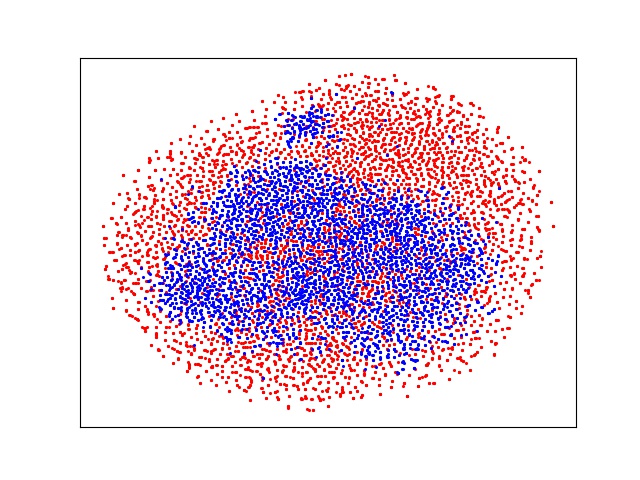}}
\subfigure[AIN (ours)]{
\label{fig3c} 
\includegraphics[width=4.3cm]{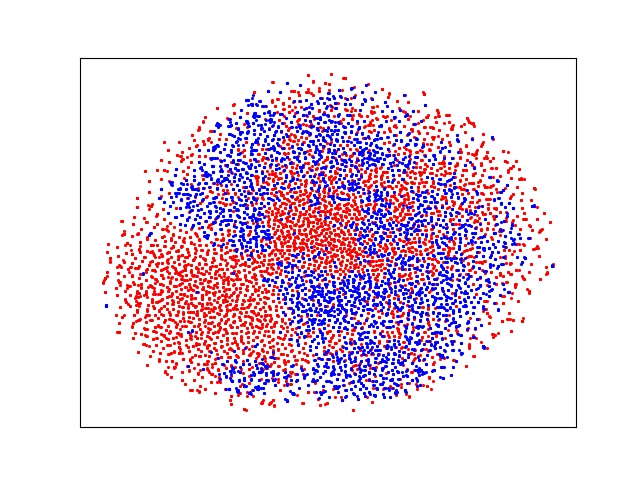}}
\caption{Feature visualization in task Tone I$\rightarrow$IV. Blue points are Tone-IV samples and red are Tone-I samples. As we can see, compared to Arcface \cite{deng2018arcface} model, our AIN indeed aligns the source features and target features to a certain extent which proves the effectiveness of our AIN on reducing inter-domain discrepancy. (Best viewed in color)}
\label{visual_align} 
\end{figure*}

\begin{figure*}
\centering
\subfigure[Arcface]{
\label{fig3a} 
\includegraphics[width=4.3cm]{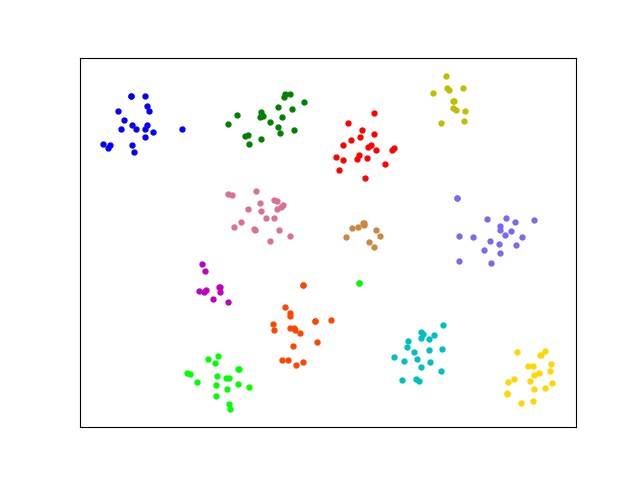}}
\hspace{0cm}
\subfigure[GCN-based clustering]{
\label{fig3b} 
\includegraphics[width=4.3cm]{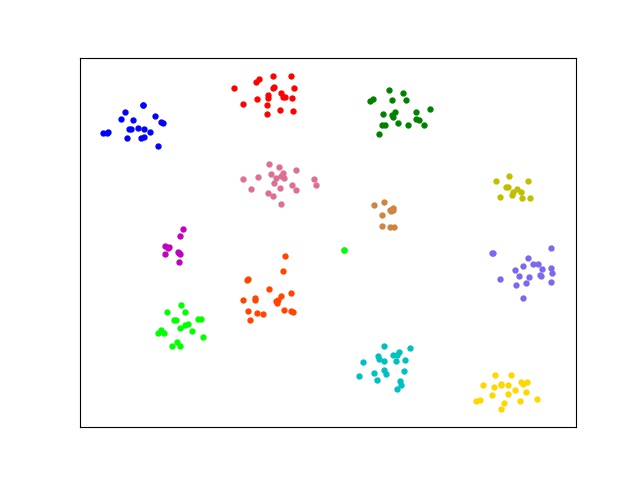}}
\subfigure[AIN (ours)]{
\label{fig3c} 
\includegraphics[width=4.3cm]{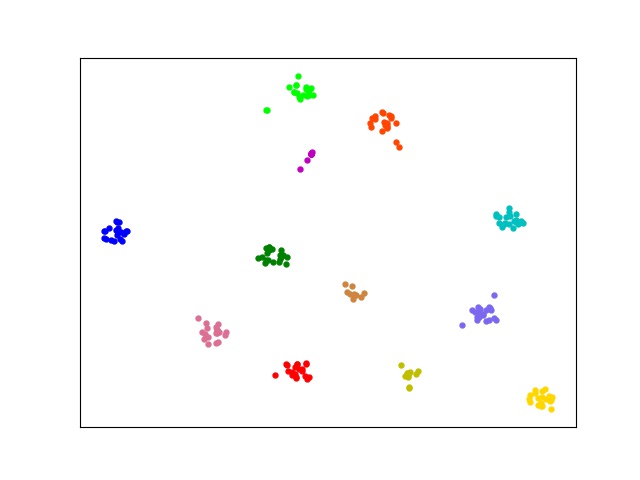}}
\caption{Feature visualization with t-SNE on target domain in task Tone I$\rightarrow$IV. The color of a circle represents its class. We observe that features of our method are more discriminative than those of other methods which proves the effectiveness of our AIN on reducing variations. (Best viewed in color)}
\label{visual_target} 
\end{figure*}

\begin{figure*}
\centering
\subfigure[Sensitivity to $k_1$]{
\label{sensitivity_k1} 
\includegraphics[width=4.3cm]{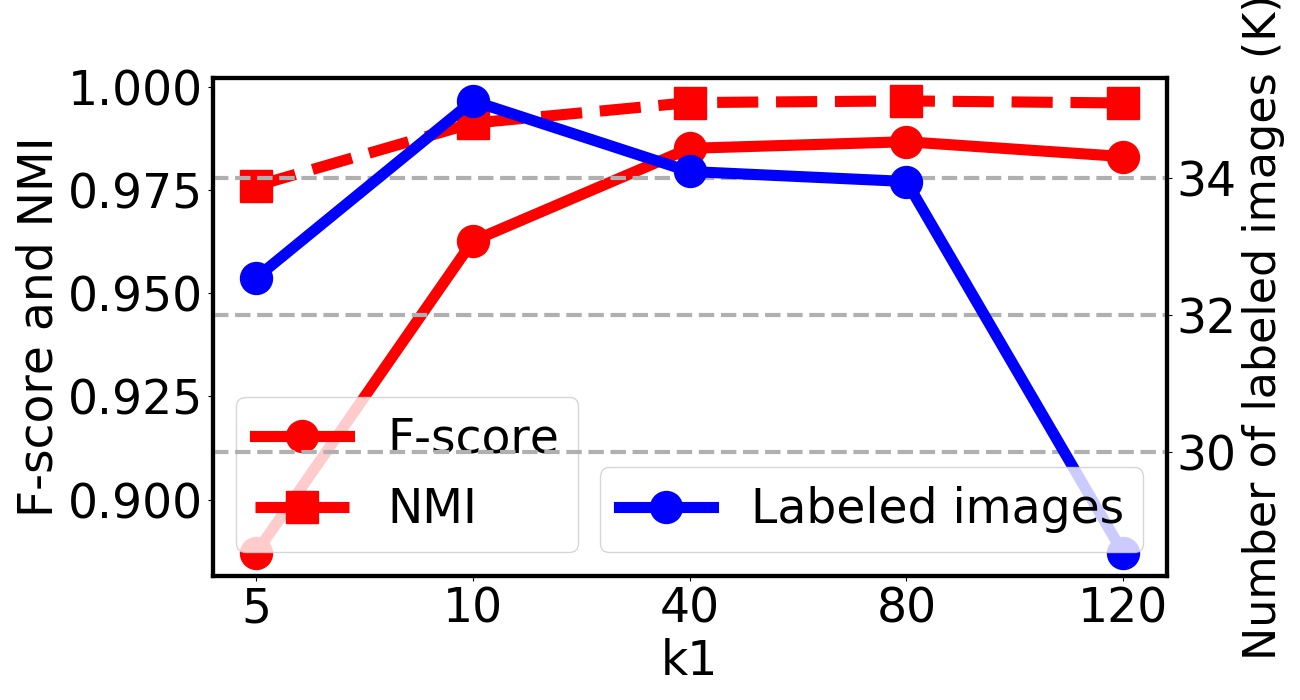}}
\hspace{0cm}
\subfigure[Sensitivity to $k_2$]{
\label{sensitivity_k2} 
\includegraphics[width=4.3cm]{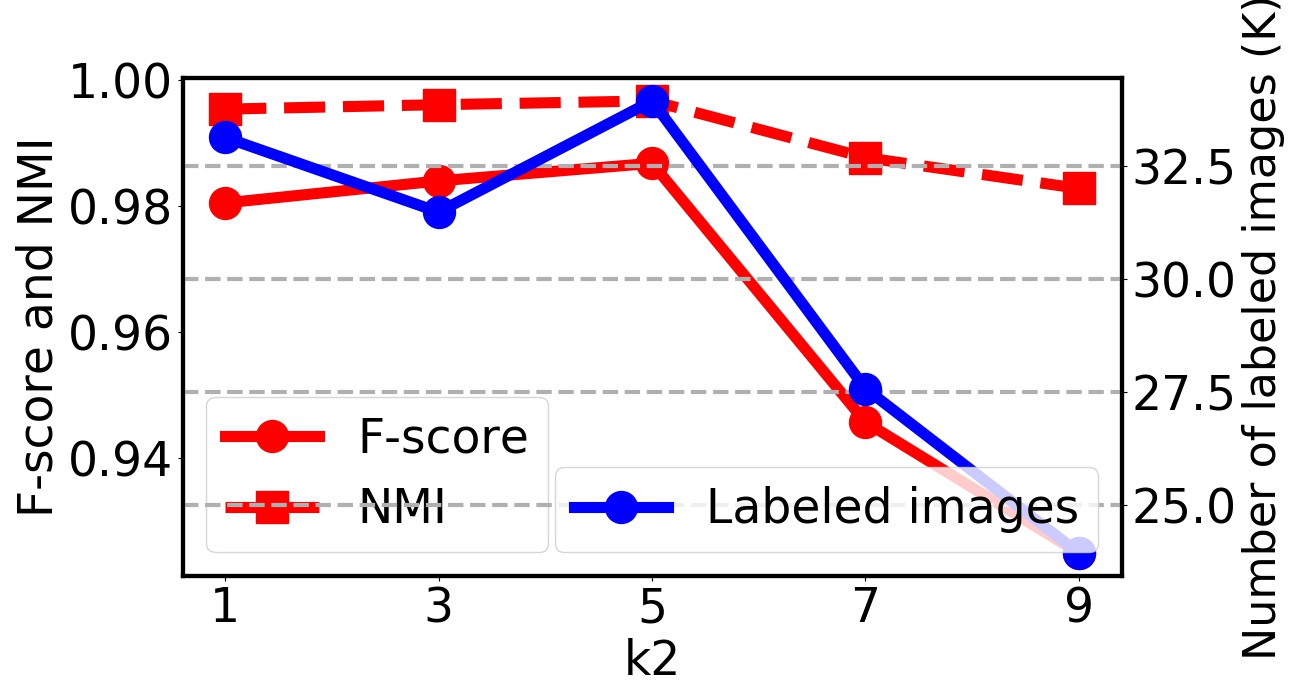}}
\subfigure[Sensitivity to $k_3$]{
\label{sensitivity_k3} 
\includegraphics[width=4.3cm]{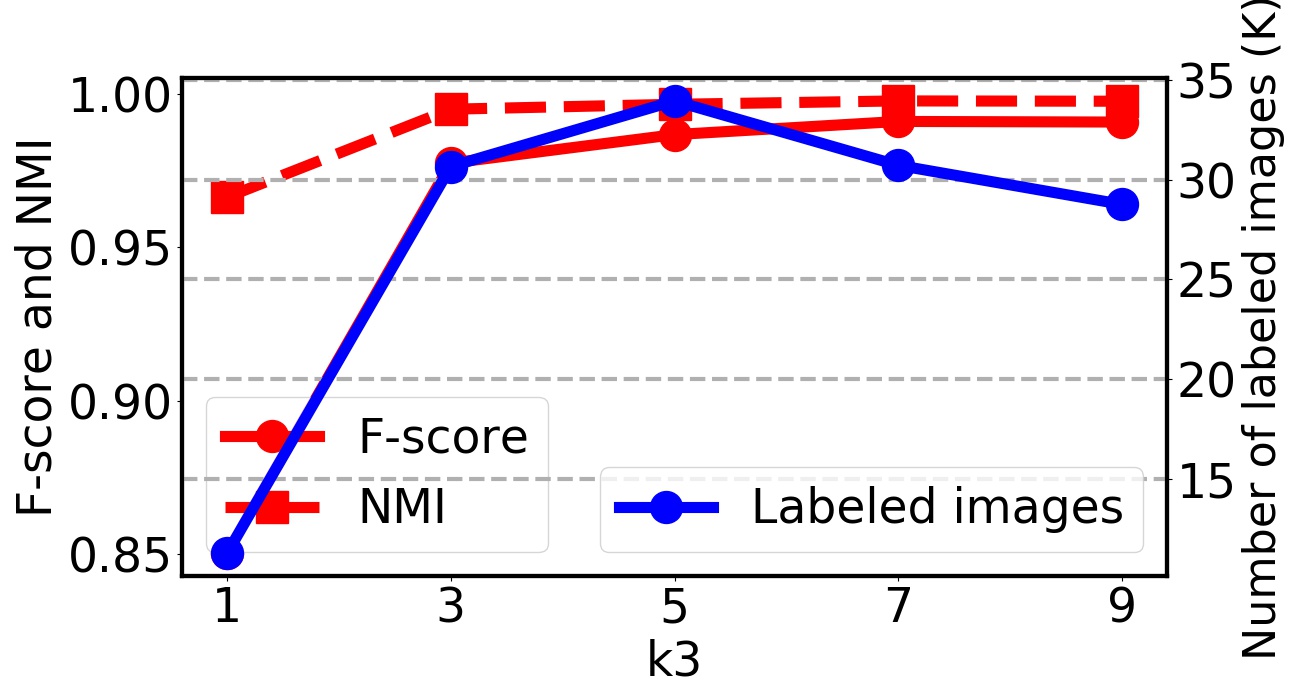}}
\subfigure[Sensitivity of accuracy to $k_2$]{
\label{sensitivity_k2_acc} 
\includegraphics[width=4.3cm]{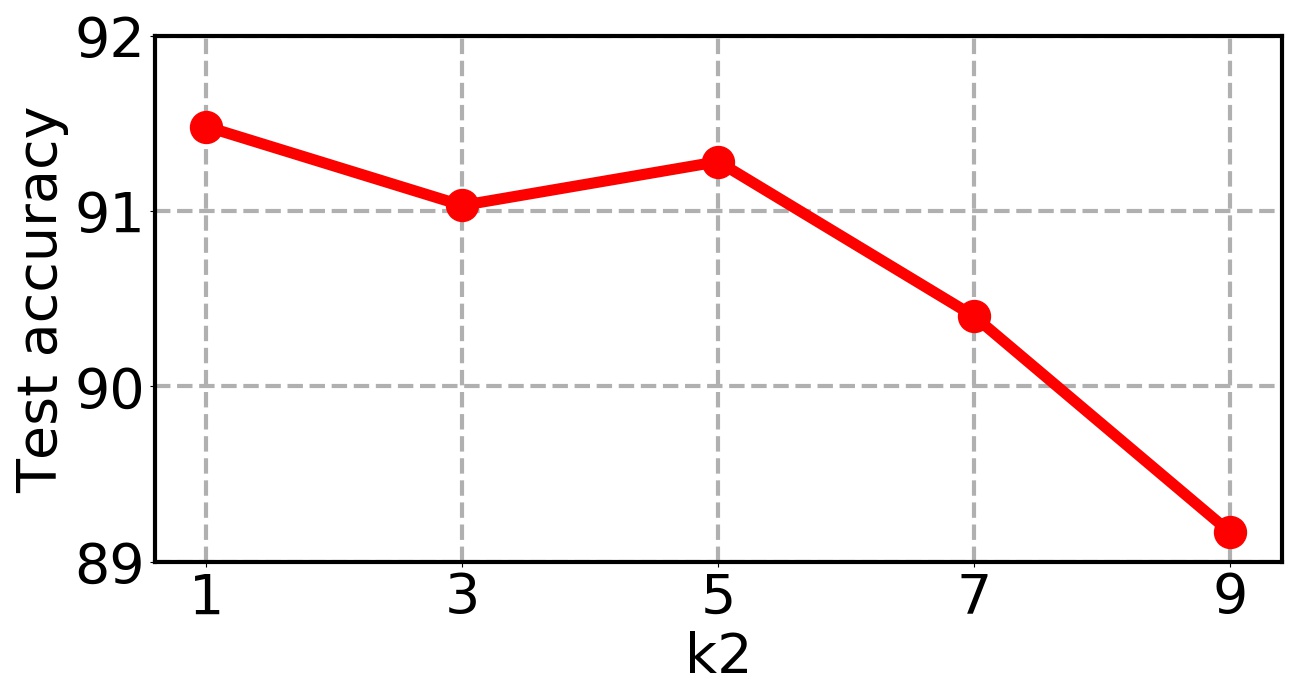}}
\caption{The sensitivity of F-score, NMI, the number of pseudo-labeled samples and target accuracy to $k_1$, $k_2$ and $k_3$ when adapting AIN-A from Tone-I to Tone-IV domain.}
\label{sensitivity_k} 
\end{figure*}

\textbf{Cooperation with other global alignment method.} Our AIN method utilizes global alignment method, i.e., MMD \cite{Tzeng2014Deep}, to reduce inter-domain discrepancy. In place of MMD, we apply another global alignment method, i.e., CORAL \cite{sun2016deep}, in our AIN-A, and evaluate its performance on RFW dataset shown in Table \ref{AIN W/ CORAL}. From the results, we observe that AIN with CORAL can also successfully improve the target performance benefiting from a good alignment between the two domains. It proves that the idea of our AIN can perform jointly with most global alignment methods.
\begin{table}[htbp]
	\begin{center}
    \caption{Verification accuracy (\%) on RFW dataset when different global alignment methods are utilized in AIN-A.}
    \label{AIN W/ CORAL}
    \small
    \setlength{\tabcolsep}{3.5mm}{
	\begin{tabular}{l|ccc}
		\hline
         Methods & II & III & IV \\ \hline \hline
         AIN-A w/ MMD & 92.47 &  93.93  & 91.28 \\
         AIN-A w/ CORAL & 92.62 &  94.10  & 91.73 \\ \hline
	\end{tabular}}
    \end{center}
\end{table}

\subsection{Visualization analysis} \label{visual experiment}

\textbf{Feature visualization.} To demonstrate the effectiveness of our AIN on reducing inter-domain discrepancy, we visualize the features on source and target domains in Fig. \ref{visual_align} using t-SNE \cite{maaten2008visualizing}. 10K source and target images are randomly chosen for better visualization and we conduct the experiment on Tone I$\rightarrow$IV task. The visualizations of features of Arcface model \cite{deng2018arcface}, GCN-A model and AIN-A model are shown, respectively. After adaptation, target features are aligned with source features so that there is no boundary between them. Moreover, we utilize MMD to compute discrepancy across domains with the features of Arcface and AIN-A. Fig. \ref{mmd_gin} shows that discrepancy on AIN-A features is much smaller than that on Arcface features, which validates that AIN-A successfully reduces inter-domain shift.

To demonstrate the effectiveness of our AIN on reducing variations within target domain, we also show feature visualization with t-SNE on target domain in task Tone I$\rightarrow$IV. The images of 12 people in Tone-IV set are randomly selected, and their features are extracted by Arcface model \cite{deng2018arcface}, GCN-A model and AIN-A model, respectively. As seen from Fig. \ref{visual_target}, features of our method are clustered tightly and show better discrimination than those of other methods. 

\begin{figure*}
\centering
\includegraphics[width=15cm]{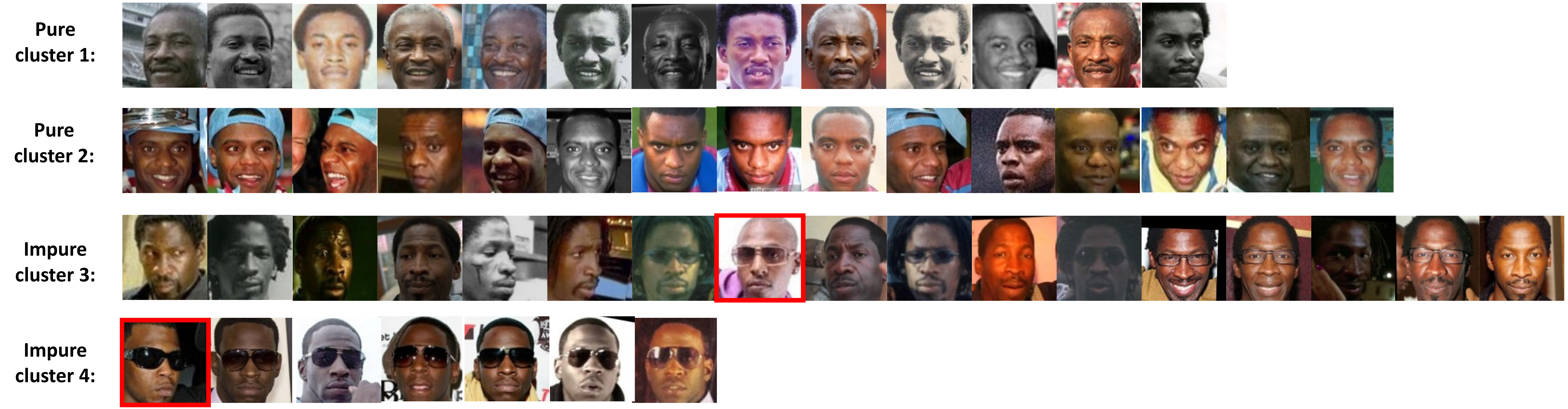}
\caption{ This figure shows four pseudo-clusters of faces generated by our method. The first two rows represent ``pure" clusters contained no noise. The last two rows represent ``impure" clusters with intra-noise, and the faces in red boxes are falsely labeled by our method. }
\label{cluster_exam}
\end{figure*}

\begin{figure}[htbp]
\centering
\includegraphics[width=8cm]{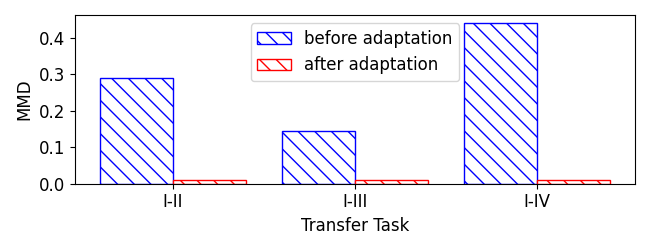}
\caption{ Discrepancy across domains with features of Arcface and AIN-A. ``I-II" means the discrepancy between Tone-I and Tone-II subjects.} 
\label{mmd_gin}
\end{figure}

\textbf{Parameter sensitivity of GCN-based clustering.} 
We perform sensitivity experiments to investigate how $k_1$, $k_2$ and $k_3$ influence F-score, NMI and the number of pseudo-labeled data in Fig. \ref{sensitivity_k1}-\ref{sensitivity_k3}. The clustering performance first increases and then decreases as $k_1$, $k_2$ and $k_3$ vary and demonstrates a bell-shaped curve. Larger $k_1$ and $k_2$ bring more candidates to be predicted, thus yield higher recall. However, more noisy neighbors would be included leading to lower precision when $k_1$ and $k_2$ become too large. Similarly, larger $k_3$ will produce more link edges and enable GCN to aggregate feature information from more neighbors; while more noisy neighbors will be aggregated when $k_3>5$. We also proved that these hyperparameters indeed influence the target accucacy in Fig. \ref{sensitivity_k2_acc}. When the clustering results are not good, the proposed adversarial MI loss would introduce more clustering noise as well leading to poor performance.

\begin{figure}[htbp]
\centering
\includegraphics[width=8cm]{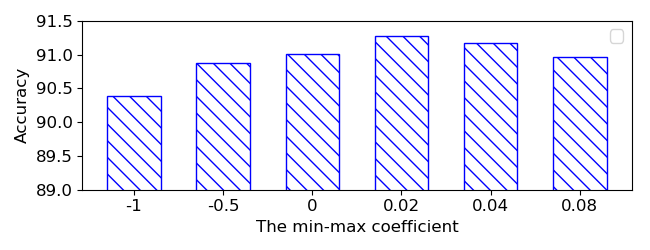}
\caption{ Target accuracy of our AIN-A influenced by the min-max coefficient $\lambda$ in task Tone I$\rightarrow$IV. Our adversarial MI will change into MI maximization used in IMAN \cite{wang2019racial} when $\lambda=-1$.}
\label{min-max coefficient}
\end{figure}

\textbf{Parameter sensitivity of adversarial MI learning.} The min-max coefficient $\lambda_1$ in Eq. \ref{min MI} will affect the adversarial learning process and so the adaptation performance. We study this parameter by setting it to different values when fixing $\lambda_2$ as 25 in Eq. \ref{max MI}, and checking the adaptation performance. Fig. \ref{min-max coefficient} shows the experimental results of AIN-A on Tone-IV subjects by using different $\lambda=\lambda_1/\lambda_2$. $\lambda>0$ denotes that $C_T$ is optimize by minimizing MI whereas $F$ is optimized by maximizing MI; $\lambda<0$ denotes that $C_T$ and $F$ is simultaneously optimized by maximizing MI. Specially, our adversarial MI will change into MI maximization used in IMAN \cite{wang2019racial} when $\lambda=-1$. As seen in Fig. \ref{min-max coefficient}, the accuracy increases as $\lambda$ increases when $\lambda<0$. That is to say, considering the intra-domain gap existed in target data, it's not beneficial when we optimize $C_T$ by maximizing MI. This will further move prototypes towards pseudo-labeled samples leading to more serious intra-domain gap. When $\lambda>0$, we observe that the accuracy first increases and then decreases as $\lambda$ varies and our AIN-A performs best when $\lambda = 0.02$. This proves that adversarial learning between $F$ and $C_T$ guided by an appropriate positive value of $\lambda$ can move prototypes towards unlabeled samples and make the optimization of unlabeled samples easy. However, a larger $\lambda$ will cause the collapse of $C_T$ leading to poor performance.

\begin{figure}
\centering
\subfigure[Target accuracy]{
\label{accuracy_iter} 
\includegraphics[width=4.1cm]{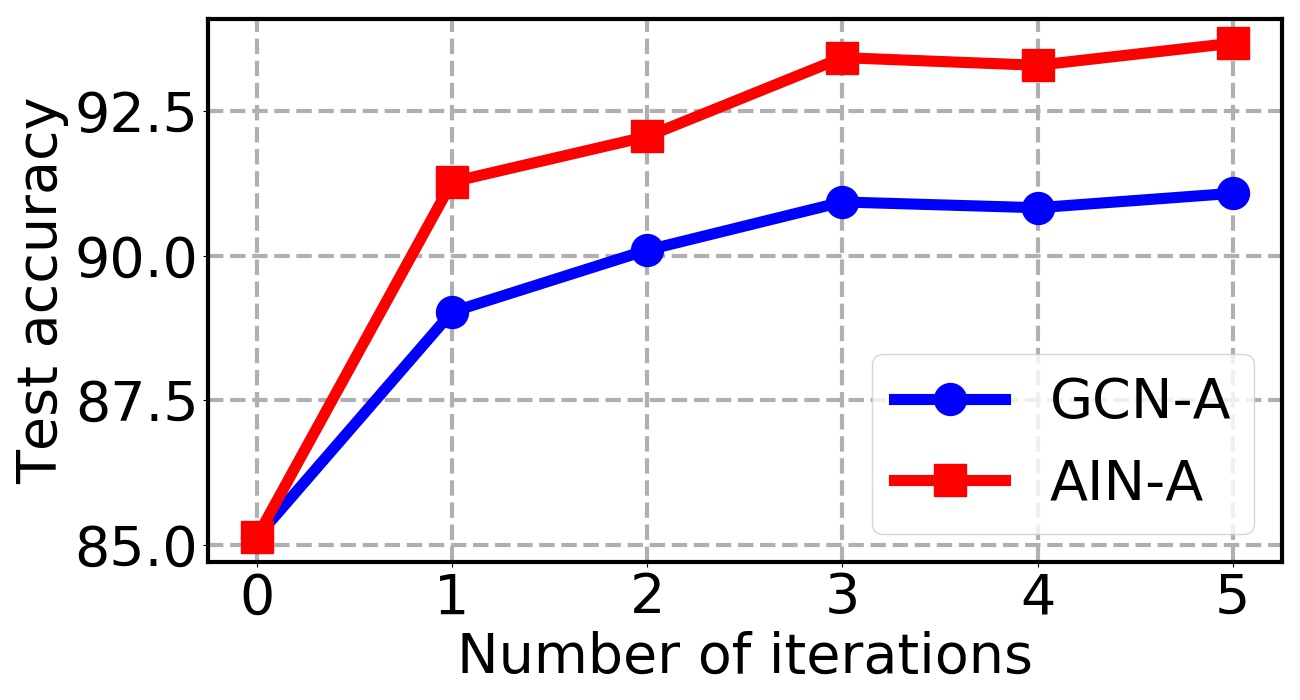}}
\hspace{0cm}
\subfigure[NMI and F-score]{
\label{NMI_iter} 
\includegraphics[width=4.1cm]{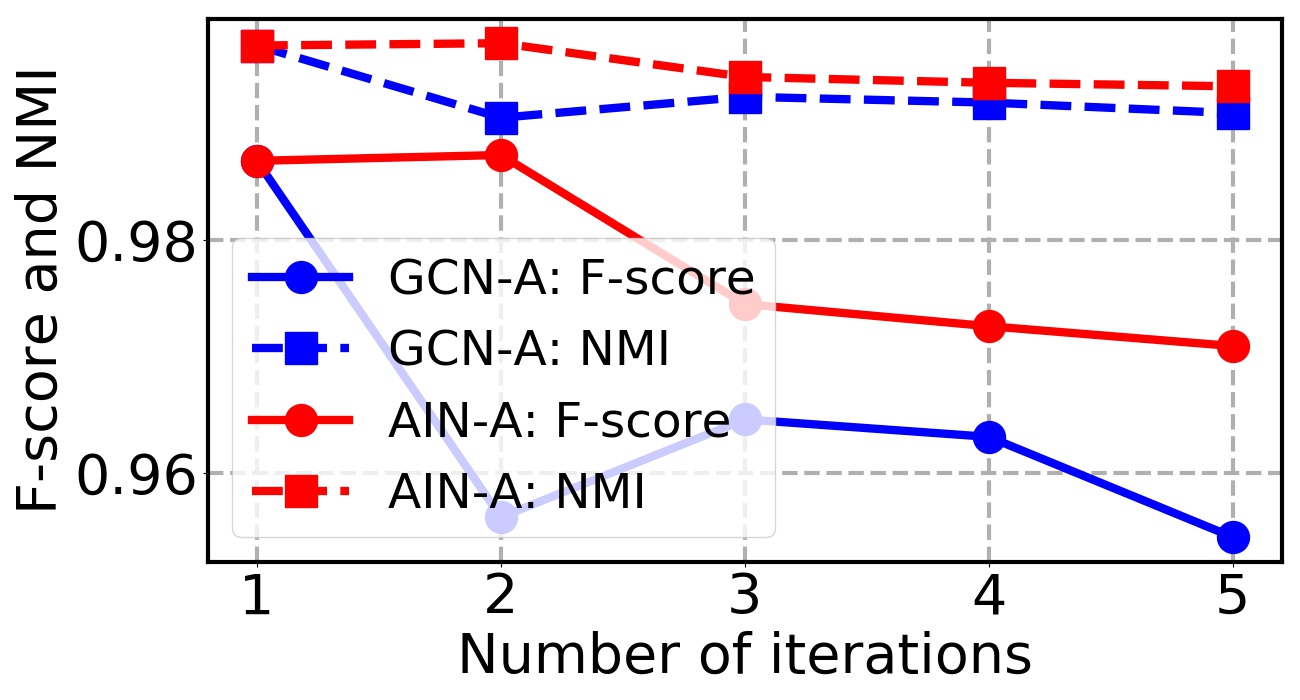}}
\subfigure[Number of pseudo-labeled data]{
\label{cluster_iter} 
\includegraphics[width=4.1cm]{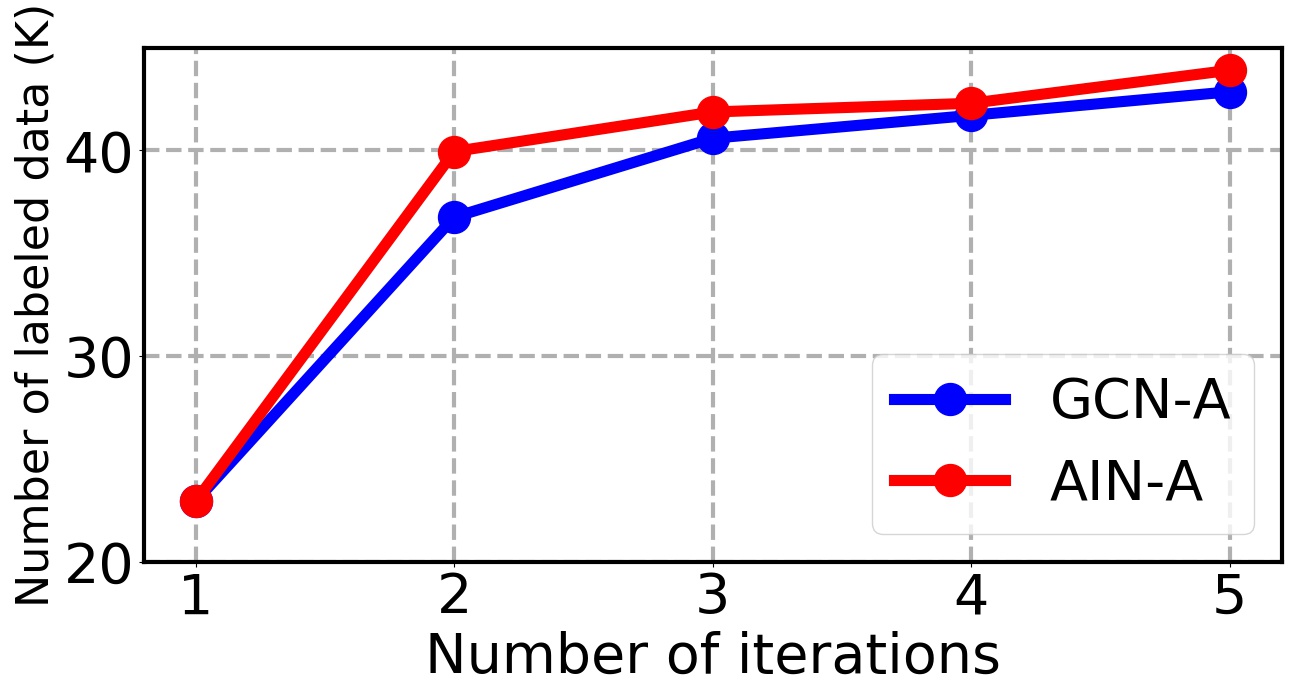}}
\hspace{0cm}
\subfigure[Number of clusters]{
\label{out_iter} 
\includegraphics[width=4.1cm]{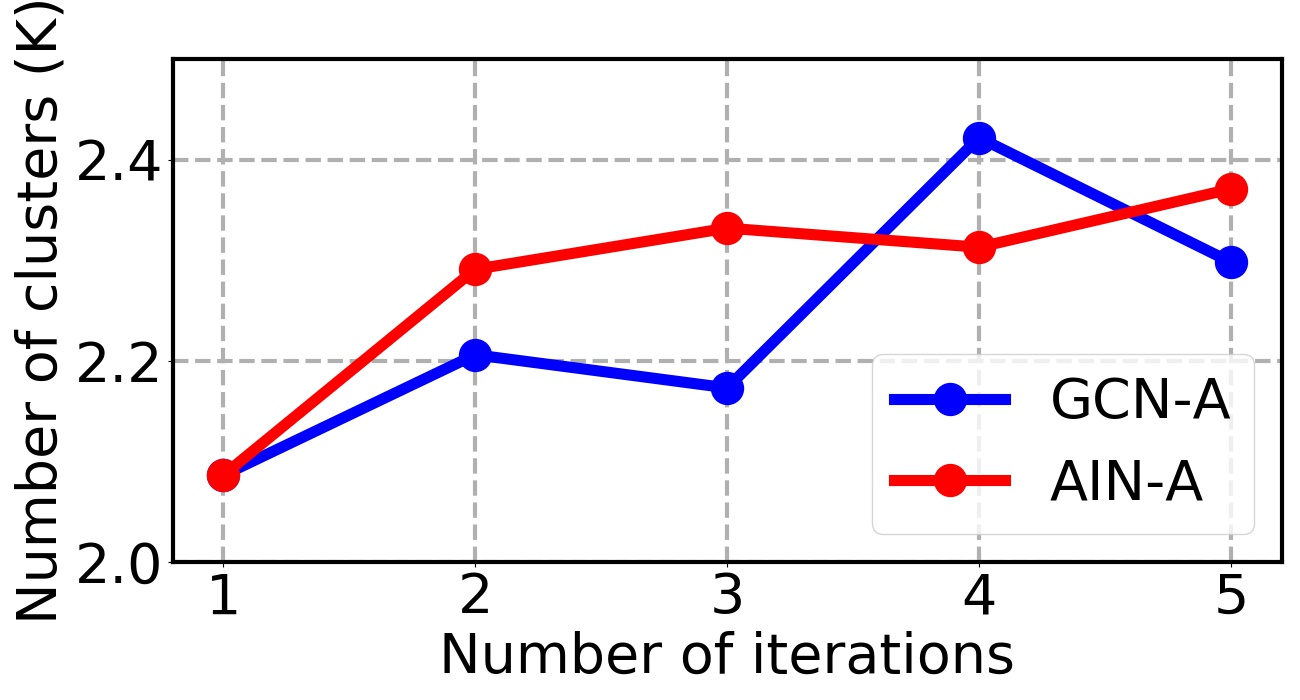}}
\caption{Performance of AIN-A at each iteration when adopting GCN-based clustering and adversarial MI learning alternatively in task Tone-I$\rightarrow$IV.}
\label{iter} 
\end{figure}

\textbf{Examples of clustering.} We visualize the clustering results of GCN in Fig. \ref{cluster_exam}. The results are obtained by AIN-A when performing the task Tone I$\rightarrow$IV. The first two rows represent ``pure" clusters contained no noise. It proves that our GCN is highly precise in identity label assignment, regardless of diverse backgrounds, expressions, poses, ages and illuminations. The last two rows represent ``impure" clusters with intra-noise, and the faces in red boxes are falsely labeled by our method. As we can see that these falsely-labeled images always have similar appearances and attributes with others in the same cluster, e.g., hairstyle and sunglasses, which are confusing even for human observers with careful inspection.

\textbf{Easy-to-hard strategy.} We additionally apply the easy-to-hard strategy of CL \cite{bengio2009curriculum} in our AIN. GCN-based clustering and adversarial MI loss are performed iteratively such that more samples can be gradually assigned pseudo labels during training. Fig. \ref{iter} illustrates the corresponding target accuracy and clustering performance during training. With the increased number of iteration, the number of pseudo-labeled data indeed increases, and the number of clusters is gradually approaching the number of ground-truth classes (2995 identities). Although more pseudo-labeled samples result in a slight decrease in BCubed F-score and NMI, they still keep higher (F-score$>$0.97 and NMI$>$0.99) throughout as training proceeds. Therefore, the iteration training of GCN-based clustering and adversarial MI loss slightly decreases the clustering performance with respect to F-score and NMI but largely boosts the clustering performance with respect to the number of clustered images. Benefitting from supervision with more pseudo-labeled data, the target accuracy gradually increases until convergence. Moreover, we can see that our AIN-A outperforms GCN-A which proves the performance can be further improved when mitigating intra-domain gap by our adversarial MI loss during curriculum learning, since our loss can make harder samples easier to learn.

\section{Conclusion}

To overcome the mismatch between domains in face recognition systems, we designed a novel AIN method. Besides dealing with inter-domain discrepancy, our AIN explicitly considered the intra-domain gap of target domain and learned discriminative target distribution under an unsupervised setting. First, GCN-based clustering utilized the relationships between node neighbors to generate more reliable pseudo labels, and adapted model with these pseudo labels such that variations within target domain can be reduced. Then, to address intra-domain gap between pseudo-labeled and unlabeled target samples and enhance the discrimination ability of network, an adversarial MI loss was proposed. Through a min-max game, it iteratively moved the class prototypes towards unlabeled target samples and clustered target samples around the updated prototypes. We empirically demonstrated the effectiveness of our AIN and set a new state of the art on RFW dataset. However, GCN also suffered from domain shift leading to falsely-labeled samples when trained it with source data and applied it on target data. Hence, one future trend is to investigate some effective methods to improve the transferability and generalization ability of GCN.

\section{Acknowledgments}

This work was supported in part by the National Natural Science Foundation of China under Grants No. 61871052 and BUPT Excellent Ph.D. Students Foundation CX2020207.

{
\bibliographystyle{IEEEtran}
\bibliography{egbib}
}

\end{document}